\pdfoutput=1
\documentclass[11pt]{article}

\usepackage[table]{xcolor}  % Required for coloring table rows

\usepackage[preprint]{acl}

\usepackage{booktabs}
\usepackage{float}
\usepackage{titlesec}
\usepackage{amssymb}
\usepackage{placeins}
\usepackage{tikz}
\usepackage{multirow}
\usepackage{times}
\usepackage{pgfplots}
\usepackage{pgfplotstable}
\pgfplotsset{compat=1.18}
\usepgfplotslibrary{colormaps}
\usepackage{algorithm}
\usepackage{algpseudocode}
\usepackage{latexsym}
\usepackage{tabularx} % Add this in the preamble
\usepackage{xurl}           % or \usepackage[hyphens]{url}
\usepackage{hyperref}
\usepackage{listings}

\lstdefinestyle{prompt}{
  basicstyle=\ttfamily\small,
  breaklines=true,
  breakatwhitespace=true,
  columns=fullflexible,
  keepspaces=true,
  showstringspaces=false,
  frame=none
}
\hypersetup{breaklinks=true}

\usepackage[T1]{fontenc}
\usepackage[utf8]{inputenc}

\usepackage{microtype}

\usepackage{inconsolata}
\usepackage{amsmath}
\usepackage{titlesec}
\usepackage{enumitem}

\usepackage{graphicx}
\usepackage{xcolor}
\title{CALM-IT: Enhancing Synthetic Motivational Interviewing Dialogue Quality with Conversational Dynamics}

\author{
 \textbf{Viet Cuong Nguyen\textsuperscript{1}},
 \textbf{Nhi Yen Nguyen\textsuperscript{1}},\\
 \textbf{Kristin A. Candan\textsuperscript{2}},
 \textbf{Mary Conlon\textsuperscript{2}},
 \textbf{Vanessa Rumie\textsuperscript{2}},
\textbf{Kristen Risola\textsuperscript{2}},\\
 \textbf{Michael L. Birnbaum\textsuperscript{3}},
 \textbf{Munmun De Choudhury\textsuperscript{1}},
\\
\\
 \textsuperscript{1}Georgia Institute of Technology,
 \textsuperscript{2}Northwell Health,
  \textsuperscript{3}Columbia University,
\\
 \small{
   \textbf{Correspondence:} \href{johnny.nguyen@gatech.edu}{johnny.nguyen@gatech.edu}
 }
}

\begin{document}
\maketitle
\begin{abstract}
Conversational dynamics describe how rapport, emotion, and goals change over a dialogue. These changes are important in therapeutic conversations, where they can influence when and how therapists intervene. However, large language model (LLM)-based synthetic therapeutic dialogue generators rarely track how these properties evolve together, which may limit the quality of longer conversations. We introduce CALM-IT, a dual-agent framework that tracks client states and therapist-side inferences to guide strategy selection and utterance generation. We evaluate synthetic Motivational Interviewing (MI) dialogues using automated Motivational Interviewing Treatment Integrity (MITI) ratings, conversational metrics, and comparisons across dialogue lengths, with ratings from four licensed psychologists providing support for key findings. Across 8,232 dialogues covering 686 grounded scenarios, four frameworks, and three target lengths, CALM-IT achieves the highest mean scores on all four MITI 4.2 global ratings and shows the smallest aggregate change in quality between short- and long-form conditions. CALM-IT also achieves the highest simulated client redirection acceptance rate (64.3\%), with its strongest redirections followed by increased change talk and reduced sustain talk. Ablation experiments show that removing conversational dynamics components reduces dialogue quality. Together, these results suggest that explicitly modeling conversational dynamics can improve the quality and consistency of long-form synthetic MI dialogues.

\end{abstract}

\section{Introduction}

Large language models (LLMs) are increasingly used for mental health support, with around one in eight U.S. young adults reporting generative AI use when feeling sad, angry, or anxious \cite{mcbain2025use}. However, these systems can respond inappropriately to suicidal ideation, reinforce delusional beliefs, or cause severe distress \cite{pichowicz2025performance,hall2025psychosis,pbs2025aipsychosis}. Research aimed at understanding and improving these systems is limited by the availability of real therapeutic conversations, which are difficult to collect and share because of privacy concerns, legal restrictions, and stigma \cite{mayer2022gdpr}. Researchers therefore often rely on short, simplified exchanges that do not capture how therapeutic conversations develop over time. Synthetic dialogue generation offers a scalable alternative, but producing realistic long-form interactions remains challenging. Although LLMs can condition their responses on dialogue history, many existing generators do not explicitly track how rapport, emotion, and conversational goals change across turns. Without these representations, they may struggle to maintain conversational progress or select appropriate interventions as an interaction unfolds.

% \begin{figure*}[t]
% \centering
% \includegraphics[width=0.9\textwidth]{Figure1.pdf}
% \caption{\textsc{CALM-IT} architecture. Simulated client states and therapist inferences evolve across state-dependent turns to guide actions, counseling strategies, and utterances.}
% \label{fig:calm-it-arch}
% \end{figure*}

\begin{table*}[t]
\centering
\small
\begin{tabular}{lcccc}
\toprule
\textbf{System} &
\textbf{Conv. Dynamics} &
\textbf{Client Grounding Source} &
\textbf{Long-form Gen.} &
\textbf{Action Selection} \\
\midrule
KMI \cite{kim2025kmi} &
$\times$ &
$Posts$ &
$\times$ &
$\surd$ \\
CAMI + STAR \cite{yang2025cami,yang2025consistent} &
$Partial$ &
$\times$ &
$\surd$ &
$\surd$ \\
CounseLLMe \cite{de2025introducing} &
$\times$ &
$\times$ &
$\times$ &
$\times$ \\
MentalChat16K \cite{xu2025mentalchat16k} &
$\times$ &
$\times$ &
$\times$ &
$\times$ \\
SQPsych \cite{vu2025roleplaying} &
$\times$ &
$Surveys$ &
$\surd$ &
$\times$ \\
CBTAlign \cite{kim2025aligning} &
$\times$ &
$\times$ &
$\surd$ &
$\times$ \\
\midrule
\textbf{CALM-IT (ours)} &
$\surd$ &
$Posts + Surveys$ &
$\surd$ &
$\surd$ \\
\bottomrule
\end{tabular}
\caption{Comparison of \texttt{CALM-IT} with prior systems generating synthetic mental health support dialogues. \texttt{CALM-IT} uniquely combines conversational dynamics tracking, cognitively grounded patient profiles, long-form multi-turn generation, and dual-actor action selection.}
\label{tab:novelty}
\end{table*}

Synthetic dialogues can also provide training and evaluation data at scale, an approach that has shown promise in domains such as instruction following and conversational modeling \cite{wang2023self,ding2023enhancing}. For therapeutic dialogue, however, the usefulness of synthetic data depends on how well it captures the interactional processes involved in counseling. Existing mental health simulation research often focuses on one participant. Client simulators commonly use relatively fixed, cooperative personas with limited resistance or emotional variation \cite{wang2024patient}, while therapist-focused systems emphasize the quality of individual responses \cite{louie2024roleplay}. These approaches do not necessarily account for how changes in one participant influence the other's behavior over time. As a result, generated conversations may move toward resolution too early, lose track of the client's motivations, or introduce interventions before the client is ready.

We introduce CALM-IT, a dual-agent framework that explicitly models conversational dynamics to generate long-form synthetic Motivational Interviewing (MI) dialogues. CALM-IT tracks changes in the simulated client's emotion, readiness for change, rapport, and conversational goals, while the therapist independently updates its inferences about these states. Both agents use these representations to select actions, choose therapeutic strategies, and generate responses. We focus on MI because effective MI requires therapists to recognize and respond to resistance and ambivalence as they emerge throughout a conversation. It also provides a framework for examining how client change talk and sustain talk develop across successive exchanges. We assess synthetic dialogue quality using measures based on Motivational Interviewing Treatment Integrity (MITI), additional conversational metrics, and comparisons across independently generated dialogue lengths. Our evaluation therefore focuses on whether explicitly modeling conversational dynamics improves synthetic dialogue quality, rather than on validating the simulated state trajectories against those observed in real therapy.

We evaluate CALM-IT by generating 8,232 dialogues covering 686 grounded scenarios, four generation frameworks, and three target lengths. Specifically, we address two research questions:
\begin{itemize}
    \item \textbf{RQ1:} Does explicit conversational dynamics modeling improve the quality of synthetic MI transcripts compared to existing frameworks?
    \item \textbf{RQ2:} Does explicit conversational dynamics modeling produce more favorable simulated client responses to therapist redirection compared to existing frameworks?
\end{itemize}

CALM-IT achieves the highest mean automated scores on all four MITI 4.2 global ratings and shows the smallest aggregate change in quality between short- and long-form dialogue conditions. Our main contributions are:
\begin{enumerate}
    \item An open-source framework for generating synthetic MI dialogues that explicitly models and leverages conversational dynamics across both client and therapist agents.
    \item An evaluation protocol combining MITI 4.2 measures, conversational metrics, and comparisons across dialogue lengths, with key automated findings corroborated by ratings from four licensed psychologists.
    \item A large-scale corpus of synthetic MI dialogues for fine-tuning, evaluating, and studying multi-turn therapeutic dialogue models.
\end{enumerate}

\section{Related Works}
\subsection{Synthetic Data for Healthcare LLMs}

Prior work has generated synthetic data for clinical settings spanning medical dialogues, EHR sequences, and clinical notes \citep{wang2024interactive,he2020meddialog,theodorou2023synthesize,yoon2023ehr,singhal2023large}, with more recent approaches extending to dual-actor multi-turn generation \citep{wang2024notechat,sviridov20253mdbench}. While some exceptions like NoteChat \citep{wang2024notechat} and 3MDBench \citep{sviridov20253mdbench} also adopt dual-actor generation pipelines, both target medical dialogue, where interactions are primarily oriented toward information gathering and diagnosis rather than relational change processes. Neither NoteChat nor 3MDBench model this joint evolution explicitly; both assign fixed roles without tracking how interactional state shifts across turns, leaving failure modes such as premature convergence and motivational drift unaddressed.

In counseling domains, efforts include empathy-labeled dialogues, multi-turn mental health exchanges, Cognitive Behavioral Therapy (CBT)-oriented datasets, and MI-focused resources \citep{lozoya2025synthetic,qiu2024smile,lee2024cactus,mahmood2025fully,jorke2025gptcoach,kim2025kmi, chandra2025reasoning}. While these approaches capture key structural and stylistic properties of therapeutic dialogue, they focus on local coherence or static datasets, leaving failure modes such as premature convergence and motivational drift invisible in the resulting generations.

\subsection{Agentic LLMs for Mental Health Support}
Research on agentic LLMs for mental health support includes simulated clients, therapist agents, and structured dialogue frameworks \citep{wang2024patient,louie2024roleplay,yang2025cami,kim2025kmi,vu2025roleplaying}. Despite these advances, prior systems share a critical limitation: fixed client personas and locally optimized therapist responses neither track how conversational state evolves across participants over time nor model how rapport, motivation, and goal alignment shift as the dialogue unfolds. In real-world counseling, these dynamics are not incidental; they are central to therapeutic change \citep{miller2013motivational,amrhein2003client,apodaca2009mechanisms}. As shown in Table~\ref{tab:novelty}, \texttt{CALM-IT} is the only framework among those compared that combines conversational dynamics tracking, cognitively grounded client profiles, long-form generation, and dual-actor action selection.
\section{The CALM-IT Framework}

\begin{algorithm}[t]
    \caption{Client Turn Generation Algorithm (Concise)}
    \label{alg:client-main}
    \begingroup
    \small
    \raggedright
    \begin{algorithmic}[1]
        \Require Dialogue history, client background and states
        \State Update the dialogue summary.
        \State Appraise the last therapist turn with system-1 thinking with 50\%
        probability. Otherwise use a neutral appraisal.
        \State Update readiness, emotion, rapport, and
        goals using the appraisal and dialogue context.
        \State Select an action using the updated states.
        \State Generate three replies using the action
        and states. Append the highest-likelihood reply.
    \end{algorithmic}
    \endgroup
\end{algorithm}

\texttt{CALM-IT} generates long-form Motivational Interviewing (MI) dialogues by tracking evolving client-side interactional states and therapist-side inferences of such states across alternating turns. We define conversational dynamics as interactional properties that evolve as participants respond to one another \citep{eshghi2012conversational}. We organize these dynamics into three theoretically grounded markers: interpersonal alignment, mental-state inference, and joint activity structure, defined below:

\begin{itemize}[leftmargin=*]
    \item \textbf{Interpersonal Alignment.} Grounded in Rapport Management Theory~\citep{spencer-oatey2000}, interpersonal alignment captures evolving relational coordination, including how individual turns strengthen or weaken rapport. It distinguishes immediate affective reactions from the accumulated therapeutic relationship.

    \item \textbf{Mental-state Inference.} Inspired by Theory of Mind research~\citep{baron-cohen1995}, mental-state inference captures one participant's provisional beliefs about another's cognitive, emotional, and motivational state, distinguishing an actor's internal state from another's necessarily incomplete inference.

    \item \textbf{Joint Activity Structure.} Motivated by theories of joint action~\citep{clark1996}, joint activity structure captures how participants establish, revise, and coordinate immediate conversational goals. Misalignment or stalled progress may require clarification or a collaborative shift in focus.
\end{itemize}

In \texttt{CALM-IT}, these markers are operationalized through explicit client-side representations of rapport, emotion, readiness for change, and short-term conversational goals, alongside the therapist's corresponding inferences. Both actors update their representations as the dialogue unfolds, with the therapist inferring client states from observable exchanges rather than accessing the client's internal states. The resulting representations condition client actions, therapist strategies, and utterance generation. Algorithms~\ref{alg:client-main} and~\ref{alg:therapist-main} summarize this alternating generation cycle.

\subsection{Client Turn Generation}

CALM-IT's client simulator maintains rapport ($\rho$), emotion ($e$), stage of change ($c$), and a short-term conversational goal ($g$). After each therapist turn, it may perform an LLM-prompted client-side system-1 evaluation ($q$), rating the therapist's response on a 5-point Likert scale based on the client's background and current state. The appraisal ranges from Very Bad to Very Good and is sampled with $z\sim\mathrm{Bernoulli}(0.5)$; otherwise, $q$ defaults to Neutral. This appraisal conditions a bounded rapport update ($\Delta\rho\in[-0.10,0.05]$), linking immediate affective reactions to accumulated interpersonal alignment. The simulator also updates readiness, emotion, and conversational goals based on recent dialogue and prior states.

Given the updated state $(\rho,e,c,g)$, the client simulator selects an interactional action, such as disclose, resist, acknowledge, or shift topics. The dialogue backbone generates three responses conditioned on the selected action and client state, and a \texttt{gemma-2-9b-it}-based reranker, fine-tuned on AnnoMI client turns, selects the highest-log-likelihood candidate \cite{wu2022anno}. Algorithm~A1 and Figures~A2--A8 of the appendix provide the expanded procedure and prompts.

\begin{algorithm}[t]
    \caption{Therapist Turn Generation Algorithm (Concise)}
    \label{alg:therapist-main}
    \begingroup
    \small
    \raggedright
    \begin{algorithmic}[1]
        \Require Dialogue history and prior client-state estimates
        \State Update the dialogue summary.
        \State Infer client background, emotion, readiness,
        goals, and rapport from observable dialogue.
        \State Check for repetition or stalled progress.
        \State Select a pivot if progress stalls. Otherwise
        select an MI-consistent strategy. Use the
        inferred states to guide either choice.
        \State Generate three replies using the strategy
        and inferred states. Append the
        highest-likelihood reply.
    \end{algorithmic}
    \endgroup
\end{algorithm}

\subsection{Therapist Turn Generation}

CALM-IT's therapist simulator updates estimates of the client's background, emotion, stage of change, rapport, and current goal ($\hat{b},\hat{e},\hat{c},\hat{\rho},\hat{g}$) after each client turn. These estimates are inferred from observable dialogue without access to the client simulator's internal states. In particular, $\hat{g}$ represents the therapist's provisional understanding of what the client wants to address. Because the client goal $g$ is hidden from the therapist, $\hat{g}$ is not compared with the simulated client-side state during generation.

The inferred states jointly condition MI-stage and strategy selection. Low $\hat{\rho}$ favors rapport-building and exploratory strategies, while planning strategies are suppressed until inferred rapport exceeds a minimum threshold and $\hat{c}$ reaches at least Contemplation. Goals involving ambivalence or focus clarification support Engaging, Focusing, or Evoking, whereas commitment and concrete next steps support Planning. Thus, $\hat{g}$ informs strategy selection without deterministically mapping to a single strategy.

A dedicated LLM call detects repetitive or unproductive patterns. Only when stagnation is detected does the therapist simulator select an autonomy-supportive pivot conditioned on inferred rapport, goal, change stage, and recent dialogue, rather than pivoting on a fixed schedule \citep{nguyen2025hanging}. For each therapist turn, the dialogue backbone generates three responses conditioned on the selected strategy and inferred client state. A \texttt{gemma-2-9b-it}-based ranker, fine-tuned on MI-coded AnnoMI therapist turns, selects the highest-log-likelihood candidate \cite{wu2022anno}. In the appendix, Algorithm~A2 and Figures~A9--A16 provide the full procedure and prompts, and Table~A1 presents a complete generation example.

\section{Experimental Setup}
We evaluate four MI dialogue-generation frameworks using DeepSeek-V3.2 as the generation backbone, with replication using OLMo-3.1-32B (Appendix~D.3). Our pipeline includes context generation, dialogue simulation, and turn-, agent-, and conversation-level evaluation of MI process fidelity, conversational quality, and length robustness (Appendix Table~C1).

\subsection{Context Generation}

We construct client contexts from publicly available Reddit mental-health disclosures \citep{brocki2023deep,sharma2018mental}. After filtering and deidentification, we cluster posts using BERTopic and manually verify the five posts closest to each topic representation, yielding 686 representative posts covering sadness, anxiety, stress, and depression. From each post, we construct an MI-grounded cognitive-motivational profile describing the client's values, needs, session goal, and motivations for change \citep{miller2013motivational}. We then simulate DASS-42 responses and match each profile to a demographically consistent respondent in the OpenPsychometrics DASS dataset \citep{openpsychometrics_dass}. DeepSeek-V3.2 \citep{liu2025deepseek} generates the client background and session goal from the matched profile. The resulting vignette, profile, background, and goal form standardized, non-clinically validated contexts shared identically across frameworks (Appendix~B).

\subsection{Dialogue Generation}

Using these contexts, we generate long-form synthetic MI dialogues with four frameworks:
\begin{enumerate}[leftmargin=*]
    \item \texttt{CAMI + STAR (C + S)} \citep{yang2025cami,yang2025consistent},
    \item \texttt{KMI} \citep{kim2025kmi},
    \item \texttt{CALM-IT} with conversational-dynamics modeling (\texttt{CI}),
    \item \texttt{CALM-IT} without conversational-dynamics modeling (\texttt{CI-NC}), removing explicit state tracking and action selection for both actors.
\end{enumerate}

We select KMI and CAMI+STAR as recent MI-specific frameworks with publicly available code and prompts supporting unrestricted long-form generation. We implement CAMI+STAR by pairing CAMI's therapist simulator with STAR's client simulator, following their original implementations. For KMI, we professionally translate therapist/client prompts and examples into English, remove Korean-specific instructions, and retain its MI forecaster and decision module. Both serve as external benchmarks with their original progression mechanisms. CI-NC is our primary baseline, with the paired \texttt{CI}--\texttt{CI-NC} comparison testing the combined contribution of explicit state tracking, action selection, and state-conditioned generation across both actors. Component ablations further examine these mechanisms (Appendix~D.4).

On 10 pilot scenarios, we informally tested System-1 evaluation probabilities of $25\%$, $50\%$, and $100\%$ and rapport-update bounds of $\{[-0.05,0.05],[-0.10,0.05],[-0.10,0.10]\}$. We selected $50\%$ and $[-0.10,0.05]$ through qualitative trajectory inspection to avoid rapport saturation and abrupt state shifts, rather than systematic hyperparameter optimization.

We select target lengths of 30, 50, and 100 speaker turns to examine sustained generation requirements, motivated by reported MI session durations \citep{magill2018meta}. Each framework and target length is applied to every simulated client, with ten additional turns reserved for session closure (Appendix~A). Length robustness is assessed across independently generated target-length conditions rather than within-dialogue trajectories. Using DeepSeek-V3.2 yields $686 \times 3 \times 4 = 8{,}232$ transcripts.

\subsection{Evaluation}
We evaluate generated dialogues at the turn, agent, and conversation levels (Appendix Table~C1), using metrics grounded in MITI, MI process theory, and conversational redirection research \citep{moyers2016motivational,miller2013motivational,apodaca2009mechanisms,moyers2009session,nguyen2024taking}. Self-consistency (within agent) and entailment (between agents) are measured using a dialogue-specific NLI model, with deviations from AnnoMI reference means characterizing under- and over-alignment \citep{hayes2007change}. We estimate therapist-initiated redirection intensity using ConvoKit with an MI-conditioned likelihood model \citep{chang2020convokit}, measuring \% Accepted Redirection and $\Delta$ Change Talk @ Greatest Redirection \citep{nguyen2024taking}. Change and sustain talk are labeled using a few-shot GPT-5.1 classifier (Appendix~C), while directionality is measured using DialoFlow \citep{li2021conversations}.

For holistic measures, including the four MITI 4.2 global ratings (Empathy, Partnership, Cultivating Change Talk, Softening Sustain Talk), we use GPT-5.1 with DeepEval \citep{Ip_deepeval_2025} for rubric-based evaluation on a 5-point Likert scale, with multi-pass aggregation to improve stability. The judge evaluates only final transcripts without access to intermediate states, applying consistent criteria across frameworks. Paired comparisons use two-sided Wilcoxon signed-rank tests with Holm-Bonferroni correction. Our objective is comparative assessment of synthetic MI process quality rather than clinical efficacy estimation. Accordingly, automated metrics support large-scale comparisons, while clinician ratings provide independent corroboration of selected framework-level differences.

Four licensed psychologists independently assess a blinded sample of 96 transcripts (1.2\%; eight per framework-length condition), following standardized annotation guidelines (Appendix Figures~C1--C3). As shown in Table~\ref{tab:human_LLM_alignment_main}, human--LLM agreement is moderate to substantial across ten rubric-based criteria (weighted Cohen's $\kappa=0.56$--$0.78$). When pooled across lengths, human and LLM evaluations yield identical aggregate framework rankings for Effectiveness, Goal Alignment, and Client Consistency (Appendix Tables~C14--C16). CALM-IT ranks first under both evaluations on these criteria, with human scores of 4.70, 4.52, and 4.78, respectively. For Effectiveness and Goal Alignment, both evaluators rank CALM-IT $>$ KMI $>$ CI-NC $>$ C+S, while for Client Consistency, both rank CALM-IT $>$ CI-NC $>$ C+S $>$ KMI. These findings corroborate key comparative patterns in automated evaluation, although the limited expert sample does not establish the reliability of every individual framework difference or validate simulated clinical trajectories. Further details appear in Appendix Sections~C.2--C.5 and Tables~C2--C16.

To examine downstream transfer utility, we fine-tune OLMo-3.1-32B on CALM-IT transcripts and compare the original and fine-tuned models as therapists within KMI, holding other settings fixed (Appendix~D.5). This evaluates whether CALM-IT-generated data can improve MI-oriented dialogue behavior outside its original generation framework.

\begin{table}[t]
\centering
\small
\renewcommand{\arraystretch}{1.15}
\begin{tabular}{p{2.7cm}c p{1.7cm}}
\toprule
\textbf{Criterion} &
\textbf{Human--LLM $\kappa_w$} &
\textbf{Interpretation} \\
\midrule
Client Consistency & 0.78 & Substantial \\
Cultivating Change Talk & 0.72 & Substantial \\
Softening Sustain Talk & 0.56 & Moderate \\
Empathy & 0.67 & Substantial \\
Partnership & 0.61 & Substantial \\
Reflection Quality & 0.77 & Substantial \\
Question Quality & 0.74 & Substantial \\
Effectiveness & 0.68 & Substantial \\
Goal Alignment & 0.73 & Substantial \\
Realignment & 0.65 & Substantial \\
\bottomrule
\end{tabular}
\caption{Human--LLM annotation reliability for rubric-based evaluations.}
\label{tab:human_LLM_alignment_main}
\end{table}

\section{Research Question 1 -- Results}

\begin{table}[t]
\centering
\small
\setlength{\tabcolsep}{2pt}
\begin{tabular}{@{}p{4.2cm}@{}cccc@{}}
\toprule
\textbf{Metric} & \textbf{KMI} & \textbf{CI-NC} & \textbf{C+S} & \textbf{CI} \\
\midrule
\multicolumn{5}{l}{\textit{Turn-level}} \\
\multicolumn{5}{l}{\textit{Client}} \\
Client Readability (\(\downarrow\)) & 5.87 & 4.68 & 5.77 & \textbf{4.45} \\
\multicolumn{5}{l}{\textit{Therapist}} \\
Reflection Quality & 4.44 & 3.71 & 2.38 & \textbf{4.61} \\
Question Quality & 4.52 & 4.39 & 2.94 & \textbf{4.97} \\
Therapist Readability (\(\downarrow\)) & 6.22 & 4.98 & 6.18 & \textbf{4.93} \\
\midrule
\multicolumn{5}{l}{\textit{Agent-level}} \\
\multicolumn{5}{l}{\textit{Client}} \\
Client Consistency & 3.76 & 4.79 & 3.42 & \textbf{4.92} \\
\multicolumn{5}{l}{\textit{Therapist}} \\
Cultivating Change Talk & 4.65 & 4.13 & 3.07 & \textbf{4.72} \\
Softening Sustain Talk & 4.73 & 4.27 & 2.53 & \textbf{4.88} \\
Empathy & 4.66 & 4.68 & 3.53 & \textbf{4.88} \\
Partnership & 4.48 & 4.60 & 2.71 & \textbf{4.88} \\
Reflection-Question Ratio & \textbf{2.05} & 0.92 & 0.90 & 1.03 \\
\% Open Questions & 97.30 & 97.90 & 96.80 & \textbf{98.47} \\
\midrule
\multicolumn{5}{l}{\textit{Conversation-level}} \\
Effectiveness & 4.27 & 3.67 & 2.14 & \textbf{4.45} \\
Realignment & 4.49 & 4.53 & 3.21 & \textbf{4.87} \\
Goal Alignment & 4.60 & 3.89 & 2.13 & \textbf{4.73} \\
\(\Delta_{\mathrm{Ref}}\) Directionality (\(\downarrow\)) & 0.081 & 0.030 & 0.048 & \textbf{0.005} \\
\(\Delta_{\mathrm{Ref}}\) Avg. Self-Consistency (\(\downarrow\)) & 0.291 & 0.141 & \textbf{0.072} & 0.157 \\
\(\Delta_{\mathrm{Ref}}\) Avg. Entailment (\(\downarrow\)) & 0.092 & 0.141 & 0.073 & \textbf{0.042} \\
\bottomrule
\end{tabular}
\caption{Turn-, agent-, and conversation-level evaluation of DeepSeek-V3.2-generated MI dialogues across KMI, CALM-IT without conversational dynamics (\texttt{CI-NC}), CAMI+STAR (\texttt{C+S}), and \texttt{CALM-IT}. Scores are averaged across scenarios and target lengths. Lower values are better for metrics marked \(\downarrow\), while higher values are better otherwise. Bold indicates the best framework.}
\label{tab:main_results}
\end{table}

\begin{figure*}[ht]
    \centering
    \makebox[\textwidth][c]{%
        \hspace{0.2cm}%
        \includegraphics[width=1.2\textwidth]{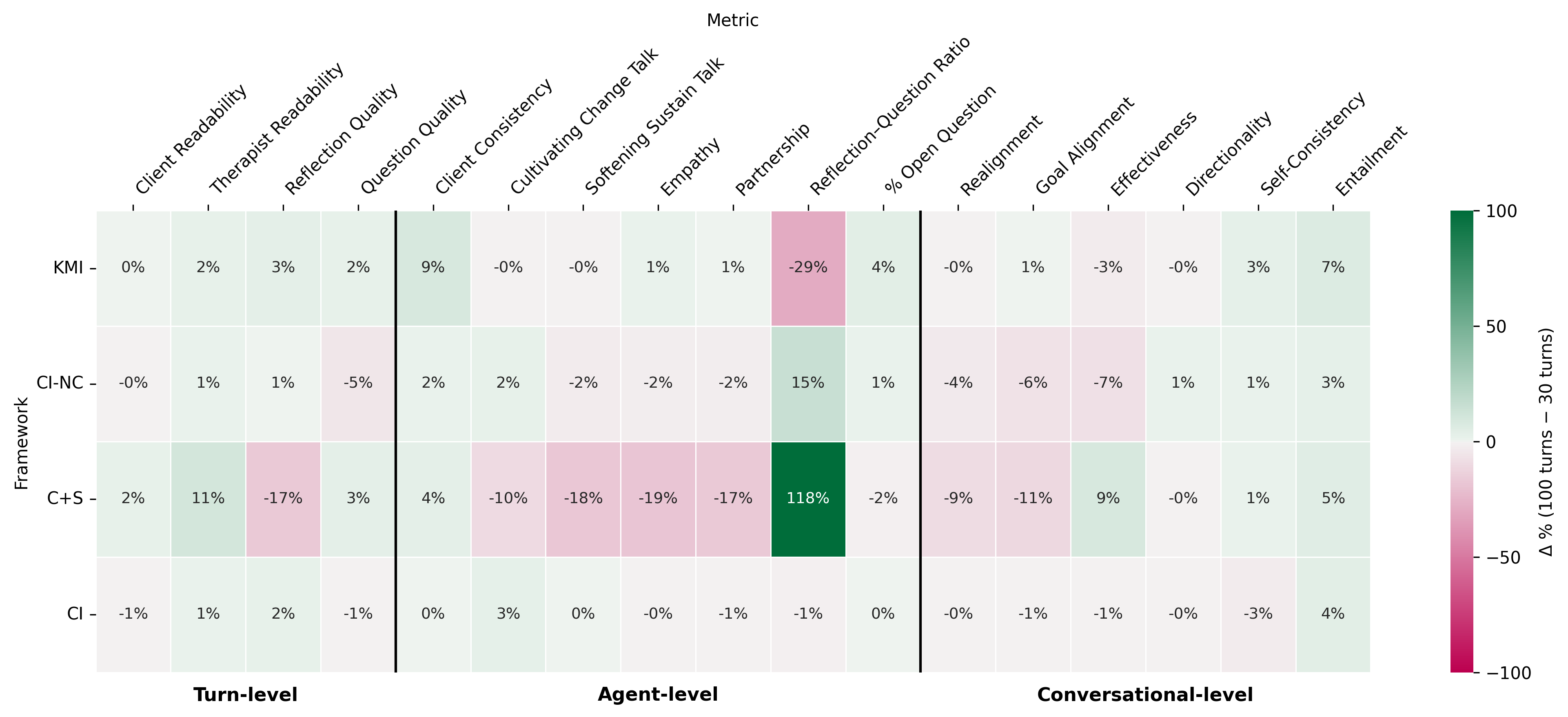}%
    }
    \caption{Percentage change between independently generated 30- and 100-turn dialogue conditions using DeepSeek-V3.2, averaged across scenario-matched comparisons. Positive values indicate increases in metric values, negative values indicate decreases, and values near zero indicate minimal change.}
    \label{fig:target_length_robustness}
\end{figure*}

Appendix Table~D1 reports descriptive statistics for all framework-length combinations. Averaged across target lengths, \texttt{CALM-IT} performs best on most turn-, agent-, and conversation-level metrics in Table~\ref{tab:main_results}. Its \textit{Effectiveness} score is \(4.45\), compared with \(4.27\) for \texttt{KMI}, \(3.67\) for \texttt{CI-NC}, and \(2.14\) for \texttt{C+S}. \texttt{CALM-IT} also achieves the highest \textit{Goal Alignment} score (\(4.73\)), significantly outperforming all alternatives (\(p<0.001\)).

% Retain the Goal Alignment significance claim only if
% supported by the corrected paired statistical tests.

It obtains the highest mean automated scores on all four MITI~4.2 global therapist ratings: \textit{Cultivating Change Talk} at \(4.72\), and \textit{Softening Sustain Talk}, \textit{Empathy}, and \textit{Partnership} each at \(4.88\). Its Cultivating Change Talk advantage over KMI (\(4.65\)) is modest at \(0.07\) points. It also achieves the best client readability (\(4.45\), lower is better) and consistency (\(4.92\), versus \(4.79\) for \texttt{CI-NC}, \(3.76\) for \texttt{KMI}, and \(3.42\) for \texttt{C+S}).

The \texttt{CI}--\texttt{CI-NC} comparison supports the combined contribution of dynamics-aware generation. \textit{Effectiveness} increases by \(0.78\) points (\(21.25\%\)), while gains on MITI globals are \(0.59\) for Cultivating Change Talk, \(0.61\) for Softening Sustain Talk, \(0.28\) for Partnership, and \(0.20\) for Empathy. Component ablations further show reduced performance when rapport tracking, goal tracking, or dynamic state updates are removed (Table~D4). Similar comparative patterns hold with OLMo-3.1-32B (Table~D3). Expert ratings also favor \texttt{CALM-IT} over \texttt{CI-NC} in Effectiveness (\(4.70\) vs.\ \(3.78\)), Goal Alignment (\(4.52\) vs.\ \(4.13\)), and Client Consistency (\(4.78\) vs.\ \(4.48\)), corroborating key aggregate differences (Table~C13).

\begin{figure*}[ht]
    \centering
    \makebox[\textwidth][c]{%
        \hspace{0.2cm}%
        \includegraphics[width=1\textwidth]{figure_3_final.png}%
    }
    \caption{Distribution of changes in client sustain talk following the strongest therapist-initiated redirection in each DeepSeek-V3.2-generated dialogue. Negative values indicate decreased sustain talk, positive values indicate increased sustain talk, solid lines show framework means, and dashed lines indicate no change.}
    \label{fig:redirection_sustain_talk}
\end{figure*}

% Length-based percentages are retained from the original analysis.
% Verify against per-length results before submission.

\texttt{CALM-IT} also exhibits greater quality robustness across target-length conditions. Figure~\ref{fig:target_length_robustness} compares independently generated 30- and 100-turn dialogues for matched scenarios. Across metrics, \texttt{CALM-IT} has the mean signed percentage change closest to zero (\(0.02\%\)), compared with \(2.41\%\) for \texttt{KMI}, \(0.72\%\) for \texttt{CI-NC}, and \(3.56\%\) for \texttt{C+S}. For conversation-level metrics, its mean change is \(-1.19\%\), compared with \(-7.05\%\) for \texttt{KMI} and \(-8.48\%\) for \texttt{CI-NC}. These results characterize aggregate quality robustness across length conditions rather than within-dialogue trajectories. 
\section{Research Question 2 -- Results}
To compare redirection-associated outcomes across dialogue-generation frameworks, we examine simulated client responses following redirection events using conversation-level metrics capturing immediate acceptance and subsequent conversational shifts. Because RQ2 concerns generated interaction trajectories, these outcomes reflect the combined behavior of each framework's therapist and client simulators rather than isolating therapist effectiveness. The \textit{percentage accepted redirection} measures the simulated client's immediate response following redirection. \texttt{CALM-IT} achieves the highest acceptance rate at \(64.28\%\), significantly outperforming \texttt{KMI} at \(59.11\%\) (\(p<0.001\)), \texttt{CAMI+STAR} at \(54.95\%\) (\(p<0.001\)), and \texttt{CI-NC} at \(52.14\%\) (\(p<0.001\)). Beyond immediate acceptance, CALM-IT's strongest redirection events are followed by mean changes of \(+12.40\%\) in client change talk and \(-8.15\%\) in sustain talk, compared with \(+9.50\%\) and \(-0.62\%\), respectively, for \texttt{KMI}.

To characterize the distributional patterns in Figure~\ref{fig:redirection_sustain_talk}, we test whether $\Delta$ sustain talk exhibits statistically significant skewness. \texttt{CALM-IT} shows significant left-skew, with skewness $\gamma=-0.66$ and a D'Agostino skewness test statistic of $Z=-6.56$ ($p<0.0001$), indicating a more pronounced tail toward larger reductions in sustain talk. The \texttt{KMI} distribution is more weakly left-skewed ($\gamma=-0.24$, $p<0.0001$), while \texttt{CAMI+STAR} and \texttt{CI-NC} exhibit significant right-skew ($\gamma=0.19$, $p<0.0001$ and $\gamma=0.66$, $p<0.0001$, respectively), indicating asymmetry toward larger increases in sustain talk. Together, acceptance rates, mean changes, and distributional patterns indicate that \texttt{CALM-IT} generates more favorable redirection-associated client responses under the evaluated simulation conditions, reflecting framework-level conversational behavior rather than independently validated therapeutic outcomes.

\section{Discussion \& Conclusion}
We introduced \textsc{CALM-IT}, a framework for generating long-form synthetic Motivational Interviewing (MI) dialogues by modeling conversational dynamics between client and therapist agents. It tracks changes in client rapport, emotion, and goals, while the therapist independently estimates these states to guide strategies and responses. Across 8,232 synthetic dialogues spanning three target lengths and four frameworks, \textsc{CALM-IT} achieves the highest mean automated scores on all four MITI 4.2 global ratings, Goal Alignment, and Effectiveness, with key comparisons supported by expert ratings. Comparing CI with CI-NC shows improvements from incorporating conversational dynamics into the shared MI-based framework, while ablations support the roles of rapport tracking, goal tracking, and dynamic state updates. \textsc{CALM-IT} also achieves higher simulated client redirection acceptance than KMI ($64.3\%$ vs.\ $59.1\%$), with its strongest redirections followed by larger reductions in sustain talk. These findings reflect the combined behavior of simulated therapist and client, not independently measured therapist effectiveness. Fine-tuning OLMo-3.1-32B on \textsc{CALM-IT} transcripts also significantly improves five of nine therapist-side MI quality measures within the KMI framework (Appendix~D.5), providing preliminary evidence that these dialogues can serve as training data outside the original framework.

Maintaining coherent conversations over extended interactions remains challenging for LLMs \citep{laban2025llms,li2025beyond}. \textsc{CALM-IT} addresses this by tracking conversational states to adjust therapeutic strategies throughout dialogue. Its relatively stable aggregate quality scores across independently generated dialogue lengths suggest less sensitivity to longer generation requirements than the evaluated baselines. However, these results compare independently generated conversations of varying lengths rather than changes within individual dialogues. They therefore cannot establish whether simulated states or trajectories accurately reflect real therapy.

Synthetic dialogue generation can supplement real therapeutic data, which remain difficult to collect and share because of privacy restrictions, limited accessibility, and stigma. It also allows systematic variation of client behaviors and interaction conditions under consistent experimental settings. Our results suggest that tracking conversational dynamics can improve synthetic MI dialogue quality and produce useful training resources, but do not establish clinical realism or therapeutic efficacy. Future work should examine within-dialogue dynamics, validate simulated state trajectories against real therapeutic interactions, and investigate how rapport, motivation, and emotion affect dialogue quality over extended conversations.
% \section{Conclusion}
% \input{latex/sections/conclusions}
\clearpage 
\section*{Limitations}
\texttt{CALM-IT} provides a framework for studying conversational dynamics in synthetic MI dialogues but does not establish the clinical realism of simulated interactional states. Real therapeutic interactions involve complex, nonlinear processes shaped by interpersonal history, situational context, and unobserved client factors, which our LLM-based state representations only partially approximate. Our study evaluates whether these representations improve generated dialogue quality rather than whether their trajectories reflect authentic therapeutic processes. Furthermore, the \texttt{CI}--\texttt{CI-NC} comparison evaluates the combined contribution of state tracking, action selection, and state-conditioned generation rather than isolating their individual effects or computational costs. Component ablations provide complementary evidence, while state-update parameters were selected through informal pilot testing rather than systematic sensitivity analysis.

A key limitation is the absence of real therapy transcripts for directly validating generated conversational trajectories. To partially address this, four licensed psychologists evaluated 96 transcripts, with human and LLM framework rankings agreeing for Effectiveness, Goal Alignment, and Client Consistency when pooled across lengths (Tables~C14--C16). This corroborates key aggregate comparisons but does not validate latent-state trajectories or establish clinical efficacy. With only eight transcripts per framework-length condition, individual comparisons have limited precision. Although generation and evaluation use different model families (DeepSeek-V3.2 and GPT-5.1), automated judgments remain sensitive to model choice, prompting, and rubric design.

Transcript generation using two backbone LLMs (DeepSeek-V3.2 and OLMo-3.1-32B) provides evidence that comparative improvements persist across model families, but broader evaluation across architectures and scales remains necessary. Similarly, downstream fine-tuning demonstrates preliminary transfer utility outside the original generation framework, without establishing superiority over alternative synthetic training corpora.

RQ2 compares complete generation frameworks with different client response policies and progression mechanisms. Its results therefore characterize joint therapist--client simulation outcomes rather than independent therapist effectiveness. Likewise, the 30-, 50-, and 100-turn conditions assess quality robustness across independently generated dialogue lengths rather than within-dialogue clinical progression or the superiority of longer sessions. Aggregate percentage changes may obscure metric-specific variation, while differences in target-length completion rates may affect comparisons.

Finally, \texttt{CALM-IT} is grounded in MI theory, limiting immediate generalization to other therapeutic modalities without corresponding changes to state representations, strategies, and evaluation criteria. Its Reddit-derived client contexts may also disproportionately reflect WEIRD (Western, Educated, Industrialized, Rich, and Democratic) populations and social-media-specific disclosure norms \citep{ma2016anonymity}. Consequently, acute crises, severe psychiatric conditions, and culturally specific concerns may be underrepresented, limiting the diversity and representativeness of generated dialogues.

\section*{Ethical Considerations}
\texttt{CALM-IT} is a research framework for studying long-form conversational dynamics in motivational interviewing, not a system intended for clinical use. The quality of the simulated conversations should not be taken as an endorsement of automated mental health support. We do not support the deployment of systems for mental health diagnosis, treatment, or crisis response without appropriate regulatory review. Clinical deployment would require appropriate safety evaluation, professional oversight, and regulatory authorization where applicable, depending on the system's intended use and risk classification. The contribution of this work lies in developing evaluation pipelines, stress-testing methods, and synthetic data resources that can support such regulatory validation efforts.

The framework relies on synthetic data to enable large-scale construction and evaluation of systems for mental health support, avoiding the privacy and legal constraints associated with real therapy transcripts. This choice also introduces limitations. Although grounded in mental health disclosures from Reddit, the resulting simulations may reflect existing biases in mental health representation, including demographic, cultural, and normative assumptions present in the source data and modeling choices. These biases can affect simulated client behavior, therapist evaluations, and assessments of therapeutic progress, potentially obscuring uneven performance across populations.

Findings obtained using \texttt{CALM-IT} should therefore be interpreted as characterizations of conversational behavior under controlled experimental conditions, not as evidence of clinical safety or effectiveness. Use of realistic mental health simulations benefits from clear boundaries, explicit documentation of data sources and assumptions, and expert oversight. In this role, synthetic frameworks such as \texttt{CALM-IT} support research, training, and regulatory preparation, while remaining distinct from the delivery of professional mental health care.

% \section*{Artifact Release and Intended Use}
% CALM-IT, generated vignettes, prompts, and evaluation protocols are intended for research, training, and stress-testing of therapeutic dialogue systems, not for clinical deployment, diagnosis, treatment, or crisis response. We will release code under MIT license and de-identified vignettes/synthetic transcripts under CC BY-NC 4.0. We do not redistribute raw Reddit posts, usernames, URLs, or direct identifiers. All released artifacts should be used only in research contexts consistent with the source-data access conditions.
% \section{Acknowledgements}
% \input{latex/sections/acknowledgments}
\bibliography{aclbib}

@article{mcbain2025use,
  title={Use of Generative AI for Mental Health Advice Among US Adolescents and Young Adults},
  author={McBain, Ryan K and Bozick, Robert and Diliberti, Melissa and Zhang, Li Ang and Zhang, Fang and Burnett, Alyssa and Kofner, Aaron and Rader, Benjamin and Breslau, Joshua and Stein, Bradley D and others},
  journal={JAMA Network Open},
  volume={8},
  number={11},
  pages={e2542281--e2542281},
  year={2025},
  publisher={American Medical Association}
}

@article{liu2025deepseek,
  title={Deepseek-v3. 2: Pushing the frontier of open large language models},
  author={Liu, Aixin and Mei, Aoxue and Lin, Bangcai and Xue, Bing and Wang, Bingxuan and Xu, Bingzheng and Wu, Bochao and Zhang, Bowei and Lin, Chaofan and Dong, Chen and others},
  journal={arXiv preprint arXiv:2512.02556},
  year={2025}
}

@inproceedings{sharma2018mental,
  title={Mental health support and its relationship to linguistic accommodation in online communities},
  author={Sharma, Eva and De Choudhury, Munmun},
  booktitle={Proceedings of the 2018 CHI conference on human factors in computing systems},
  pages={1--13},
  year={2018}
}

@article{moyers2016motivational,
  title={The motivational interviewing treatment integrity code (MITI 4): rationale, preliminary reliability and validity},
  author={Moyers, Theresa B and Rowell, Lauren N and Manuel, Jennifer K and Ernst, Denise and Houck, Jon M},
  journal={Journal of substance abuse treatment},
  volume={65},
  pages={36--42},
  year={2016},
  publisher={Elsevier}
}

@article{moyers2009session,
  title={From in-session behaviors to drinking outcomes: a causal chain for motivational interviewing.},
  author={Moyers, Theresa B and Martin, Tim and Houck, Jon M and Christopher, Paulette J and Tonigan, J Scott},
  journal={Journal of consulting and clinical psychology},
  volume={77},
  number={6},
  pages={1113},
  year={2009},
  publisher={American Psychological Association}
}

@misc{Ip_deepeval_2025,
    author = {Ip, Jeffrey and Vongthongsri, Kritin},
    license = {Apache-2.0},
    month = oct,
    title = {{deepeval}},
    url = {https://github.com/confident-ai/deepeval},
    version = {3.7.4},
    year = {2025}
}

@article{pichowicz2025performance,
  title={Performance of mental health chatbot agents in detecting and managing suicidal ideation},
  author={Pichowicz, Wojciech and Kotas, Michal and Piotrowski, Pawel},
  journal={Scientific Reports},
  volume={15},
  number={1},
  pages={31652},
  year={2025},
  publisher={Nature Publishing Group UK London}
}

@article{nguyen2024taking,
  title={Taking a turn for the better: Conversation redirection throughout the course of mental-health therapy},
  author={Nguyen, Vivian and Jung, Sang Min and Lee, Lillian and Hull, Thomas D and Danescu-Niculescu-Mizil, Cristian},
  journal={arXiv preprint arXiv:2410.07147},
  year={2024}
}

@article{dale1948formula,
  title={A formula for predicting readability: Instructions},
  author={Dale, Edgar and Chall, Jeanne S},
  journal={Educational research bulletin},
  pages={37--54},
  year={1948},
  publisher={JSTOR}
}

@misc{openpsychometrics_dass,
  title        = {Depression Anxiety Stress Scales (DASS) Dataset},
  author       = {{Open Psychometrics}},
  year         = {2019},
  howpublished = {\url{https://openpsychometrics.org/_rawdata/DASS_data_21.02.19.zip}}}

@inproceedings{ma2016anonymity,
  title={Anonymity, intimacy and self-disclosure in social media},
  author={Ma, Xiao and Hancock, Jeff and Naaman, Mor},
  booktitle={Proceedings of the 2016 CHI conference on human factors in computing systems},
  pages={3857--3869},
  year={2016}
}

@article{hayes2007change,
  title={Change is not always linear: The study of nonlinear and discontinuous patterns of change in psychotherapy},
  author={Hayes, Adele M and Laurenceau, Jean-Philippe and Feldman, Greg and Strauss, Jennifer L and Cardaciotto, LeeAnn},
  journal={Clinical psychology review},
  volume={27},
  number={6},
  pages={715--723},
  year={2007},
  publisher={Elsevier}
}

@article{laban2025llms,
  title={Llms get lost in multi-turn conversation},
  author={Laban, Philippe and Hayashi, Hiroaki and Zhou, Yingbo and Neville, Jennifer},
  journal={arXiv preprint arXiv:2505.06120},
  year={2025}
}

@article{li2025beyond,
  title={Beyond single-turn: A survey on multi-turn interactions with large language models},
  author={Li, Yubo and Shen, Xiaobin and Yao, Xinyu and Ding, Xueying and Miao, Yidi and Krishnan, Ramayya and Padman, Rema},
  journal={arXiv preprint arXiv:2504.04717},
  year={2025}
}

@inproceedings{brocki2023deep,
  title={Deep learning mental health dialogue system},
  author={Brocki, Lennart and Dyer, George C and G{\l}adka, Anna and Chung, Neo Christopher},
  booktitle={2023 IEEE International Conference on Big Data and Smart Computing (BigComp)},
  pages={395--398},
  year={2023},
  organization={IEEE}
}

@article{de2025introducing,
  title={Introducing CounseLLMe: A dataset of simulated mental health dialogues for comparing LLMs like Haiku, LLaMAntino and ChatGPT against humans},
  author={De Duro, Edoardo Sebastiano and Improta, Riccardo and Stella, Massimo},
  journal={Emerging Trends in Drugs, Addictions, and Health},
  volume={5},
  pages={100170},
  year={2025},
  publisher={Elsevier}
}

@article{nguyen2025hanging,
  title={Hanging in the Balance: Pivotal Moments in Crisis Counseling Conversations},
  author={Nguyen, Vivian and Lee, Lillian and Danescu-Niculescu-Mizil, Cristian},
  journal={arXiv preprint arXiv:2506.03941},
  year={2025}
}

@inproceedings{xu2025mentalchat16k,
  title={Mentalchat16k: A benchmark dataset for conversational mental health assistance},
  author={Xu, Jia and Wei, Tianyi and Hou, Bojian and Orzechowski, Patryk and Yang, Shu and Jin, Ruochen and Paulbeck, Rachael and Wagenaar, Joost and Demiris, George and Shen, Li},
  booktitle={Proceedings of the 31st ACM SIGKDD Conference on Knowledge Discovery and Data Mining V. 2},
  pages={5367--5378},
  year={2025}
}

@misc{hall2025psychosis,
  title        = {AI Chatbots Giving Dangerous Advice to Mentally Ill, Psychologists Warn},
  author       = {Hall, Rachel},
  year         = {2025},
  howpublished = {\url{https://www.theguardian.com/technology/2025/nov/30/chatgpt-dangerous-advice-mentally-ill-psychologists-openai}},
  note         = {The Guardian. Accessed: 2025-12-01}
}

@misc{pbs2025aipsychosis,
  title        = {What to Know About ``AI Psychosis'' and the Effect of AI Chatbots on Mental Health},
  author       = {{PBS NewsHour}},
  year         = {2025},
  howpublished = {\url{https://www.pbs.org/newshour/show/what-to-know-about-ai-psychosis-and-the-effect-of-ai-chatbots-on-mental-health}},
  note         = {Accessed: 2025-08-15}
}

@article{louie2024roleplay,
  title={Roleplay-doh: Enabling domain-experts to create llm-simulated patients via eliciting and adhering to principles},
  author={Louie, Ryan and Nandi, Ananjan and Fang, William and Chang, Cheng and Brunskill, Emma and Yang, Diyi},
  journal={arXiv preprint arXiv:2407.00870},
  year={2024}
}

@article{wang2024patient,
  title={Patient-\(\Psi\): Using large language models to simulate patients for training mental health professionals},
  author={Wang, Ruiyi and Milani, Stephanie and Chiu, Jamie C and Zhi, Jiayin and Eack, Shaun M and Labrum, Travis and Murphy, Samuel M and Jones, Nev and Hardy, Kate and Shen, Hong and others},
  journal={arXiv preprint arXiv:2405.19660},
  year={2024}
}

@article{kim2025kmi,
  title={KMI: A Dataset of Korean Motivational Interviewing Dialogues for Psychotherapy},
  author={Kim, Hyunjong and Lee, Suyeon and Cho, Yeongjae and Ryu, Eunseo and Jo, Yohan and Seong, Suran and Cho, Sungzoon},
  journal={arXiv preprint arXiv:2502.05651},
  year={2025}
}

@inproceedings{yang2025cami,
  title={Cami: A counselor agent supporting motivational interviewing through state inference and topic exploration},
  author={Yang, Yizhe and Achananuparp, Palakorn and Huang, He-Yan and Jiang, Jing and Kit, Phey Ling and Lim, Nicholas Gabriel and Ern, Cameron Tan Shi and Lim, Ee-Peng},
  booktitle={Proceedings of the 63rd Annual Meeting of the Association for Computational Linguistics (Volume 1: Long Papers)},
  pages={21037--21081},
  year={2025}
}

@article{wang2024interactive,
  title={Interactive computer-aided diagnosis on medical image using large language models},
  author={Wang, Sheng and Zhao, Zihao and Ouyang, Xi and Liu, Tianming and Wang, Qian and Shen, Dinggang},
  journal={Communications Engineering},
  volume={3},
  number={1},
  pages={133},
  year={2024},
  publisher={Nature Publishing Group UK London}
}

@article{he2020meddialog,
  title={Meddialog: Two large-scale medical dialogue datasets},
  author={He, Xuehai and Chen, Shu and Ju, Zeqian and Dong, Xiangyu and Fang, Hongchao and Wang, Sicheng and Yang, Yue and Zeng, Jiaqi and Zhang, Ruisi and Zhang, Ruoyu and others},
  journal={arXiv preprint arXiv:2004.03329},
  year={2020}
}

@article{yoon2023ehr,
  title={EHR-Safe: generating high-fidelity and privacy-preserving synthetic electronic health records},
  author={Yoon, Jinsung and Mizrahi, Michel and Ghalaty, Nahid Farhady and Jarvinen, Thomas and Ravi, Ashwin S and Brune, Peter and Kong, Fanyu and Anderson, Dave and Lee, George and Meir, Arie and others},
  journal={NPJ digital medicine},
  volume={6},
  number={1},
  pages={141},
  year={2023},
  publisher={Nature Publishing Group UK London}
}

@article{theodorou2023synthesize,
  title={Synthesize high-dimensional longitudinal electronic health records via hierarchical autoregressive language model},
  author={Theodorou, Brandon and Xiao, Cao and Sun, Jimeng},
  journal={Nature communications},
  volume={14},
  number={1},
  pages={5305},
  year={2023},
  publisher={Nature Publishing Group UK London}
}

@article{singhal2023large,
  title={Large language models encode clinical knowledge},
  author={Singhal, Karan and Azizi, Shekoofeh and Tu, Tao and Mahdavi, S Sara and Wei, Jason and Chung, Hyung Won and Scales, Nathan and Tanwani, Ajay and Cole-Lewis, Heather and Pfohl, Stephen and others},
  journal={Nature},
  volume={620},
  number={7972},
  pages={172--180},
  year={2023},
  publisher={Nature Publishing Group}
}

@article{mayer2022gdpr,
  title={GDPR compliant collection of Therapist-Patient-Dialogues},
  author={Mayer, Tobias and Warikoo, Neha and Grimm, Oliver and Reif, Andreas and Gurevych, Iryna},
  journal={arXiv preprint arXiv:2211.12360},
  year={2022}
}

@inproceedings{lozoya2025synthetic,
  title={Synthetic Empathy: Generating and Evaluating Artificial Psychotherapy Dialogues to Detect Empathy in Counseling Sessions},
  author={Lozoya, Daniel Cabrera and L{\'u}a, Eloy Hern{\'a}ndez and Perches, Juan Alberto Barajas and Conway, Mike and D’Alfonso, Simon},
  booktitle={Proceedings of the 10th Workshop on Computational Linguistics and Clinical Psychology (CLPsych 2025)},
  pages={157--171},
  year={2025}
}

@inproceedings{qiu2024smile,
  title={Smile: Single-turn to multi-turn inclusive language expansion via chatgpt for mental health support},
  author={Qiu, Huachuan and He, Hongliang and Zhang, Shuai and Li, Anqi and Lan, Zhenzhong},
  booktitle={Findings of the Association for Computational Linguistics: EMNLP 2024},
  pages={615--636},
  year={2024}
}

@inproceedings{mahmood2025fully,
  title={A Fully Generative Motivational Interviewing Counsellor Chatbot for Moving Smokers Towards the Decision to Quit},
  author={Mahmood, Zafarullah and Ali, Soliman and Zhu, Jiading and Abdelwahab, Mohamed and Collins, Michelle Yu and Chen, Sihan and Zhao, Yi Cheng and Wolff, Jodi and Melamed, Osnat C and Minian, Nadia and others},
  booktitle={Findings of the Association for Computational Linguistics: ACL 2025},
  pages={25008--25043},
  year={2025}
}

@inproceedings{lee2024cactus,
  title={Cactus: Towards psychological counseling conversations using cognitive behavioral theory},
  author={Lee, Suyeon and Mac Kim, Sunghwan and Kim, Minju and Kang, Dongjin and Yang, Dongil and Kim, Harim and Kang, Minseok and Jung, Dayi and Kim, Min Hee and Lee, Seungbeen and others},
  booktitle={Findings of the Association for Computational Linguistics: EMNLP 2024},
  pages={14245--14274},
  year={2024}
}

@inproceedings{jorke2025gptcoach,
  title={GPTCoach: Towards LLM-Based Physical Activity Coaching},
  author={J{\"o}rke, Matthew and Sapkota, Shardul and Warkenthien, Lyndsea and Vainio, Niklas and Schmiedmayer, Paul and Brunskill, Emma and Landay, James A},
  booktitle={Proceedings of the 2025 CHI Conference on Human Factors in Computing Systems},
  pages={1--46},
  year={2025}
}

@article{amrhein2003client,
  title={Client commitment language during motivational interviewing predicts drug use outcomes.},
  author={Amrhein, Paul C and Miller, William R and Yahne, Carolina E and Palmer, Michael and Fulcher, Laura},
  journal={Journal of consulting and clinical psychology},
  volume={71},
  number={5},
  pages={862},
  year={2003},
  publisher={American Psychological Association}
}

@article{eshghi2012conversational,
  title={Conversational interactions: Capturing dialogue dynamics},
  author={Eshghi, Arash and Hough, Julian and Purver, Matthew and Kempson, Ruth and Gregoromichelaki, Eleni},
  journal={From Quantification to Conversation: Festschrift for Robin Cooper on the occasion of his 65th birthday},
  volume={19},
  pages={325--349},
  year={2012}
}

@article{apodaca2009mechanisms,
  title={Mechanisms of change in motivational interviewing: A review and preliminary evaluation of the evidence},
  author={Apodaca, Timothy R and Longabaugh, Richard},
  journal={Addiction},
  volume={104},
  number={5},
  pages={705--715},
  year={2009},
  publisher={Wiley Online Library}
}

@book{miller2013motivational,
  title     = {Motivational Interviewing: Helping People Change},
  author    = {Miller, William R. and Rollnick, Stephen},
  publisher = {Guilford Press},
  edition   = {3rd},
  year      = {2013}
}

@inproceedings{yang2025consistent,
  title={Consistent client simulation for motivational interviewing-based counseling},
  author={Yang, Yizhe and Achananuparp, Palakorn and Huang, He-Yan and Jiang, Jing and Lim, Nicholas Gabriel and Ern, Cameron Tan Shi and Kit, Phey Ling and Xiuhui, Jenny Giam and Pinto, John and Lim, Ee-peng},
  booktitle={Proceedings of the 63rd Annual Meeting of the Association for Computational Linguistics (Volume 1: Long Papers)},
  pages={20959--20998},
  year={2025}
}

@inproceedings{chang2020convokit,
  title={Convokit: A toolkit for the analysis of conversations},
  author={Chang, Jonathan P and Chiam, Caleb and Fu, Liye and Wang, Andrew and Zhang, Justine and Danescu-Niculescu-Mizil, Cristian},
  booktitle={Proceedings of the 21th annual meeting of the special interest group on discourse and dialogue},
  pages={57--60},
  year={2020}
}

@article{kim2025aligning,
  title={Aligning large language models for cognitive behavioral therapy: a proof-of-concept study},
  author={Kim, Yejin and Choi, Chi-Hyun and Cho, Selin and Sohn, Jy-yong and Kim, Byung-Hoon},
  journal={Frontiers in Psychiatry},
  volume={16},
  pages={1583739},
  year={2025},
  publisher={Frontiers}
}

@article{vu2025roleplaying,
  title={Roleplaying with Structure: Synthetic Therapist-Client Conversation Generation from Questionnaires},
  author={Vu, Doan Nam Long and Tan, Rui and Moench, Lena and Francke, Svenja Jule and Woiwod, Daniel and Thomas-Odenthal, Florian and Stroth, Sanna and Kircher, Tilo and Hermann, Christiane and Dannlowski, Udo and others},
  journal={arXiv preprint arXiv:2510.25384},
  year={2025}
}

@book{spencer-oatey2000,
  title={Culturally speaking: Managing rapport through talk across cultures},
  author={Spencer-Oatey, Helen},
  year={2004},
  publisher={A\&C Black}
}

@article{baron-cohen1995,
  title={Mindblindness: An Essay on Autism and Theory of Mind},
  author={Baron-Cohen, Simon},
  journal={MIT Press},
  year={1995}
}

@book{clark1996,
  title={Using language},
  author={Clark, Herbert H},
  year={1996},
  publisher={Cambridge university press}
}

@article{li2021conversations,
  title={Conversations are not flat: Modeling the dynamic information flow across dialogue utterances},
  author={Li, Zekang and Zhang, Jinchao and Fei, Zhengcong and Feng, Yang and Zhou, Jie},
  journal={arXiv preprint arXiv:2106.02227},
  year={2021}
}

@inproceedings{dziri2019evaluating,
  title={Evaluating coherence in dialogue systems using entailment},
  author={Dziri, Nouha and Kamalloo, Ehsan and Mathewson, Kory and Zaiane, Osmar R},
  booktitle={Proceedings of the 2019 conference of the north american chapter of the association for computational linguistics: Human language technologies, volume 1 (long and short papers)},
  pages={3806--3812},
  year={2019}
}

@article{chandra2025reasoning,
  title={Reasoning Is Not All You Need: Examining LLMs for Multi-Turn Mental Health Conversations},
  author={Chandra, Mohit and Sriraman, Siddharth and Khanuja, Harneet Singh and Jin, Yiqiao and De Choudhury, Munmun},
  journal={arXiv preprint arXiv:2505.20201},
  year={2025}
}

@inproceedings{wang2024notechat,
  title={Notechat: a dataset of synthetic patient-physician conversations conditioned on clinical notes},
  author={Wang, Junda and Yao, Zonghai and Yang, Zhichao and Zhou, Huixue and Li, Rumeng and Wang, Xun and Xu, Yucheng and Yu, Hong},
  booktitle={Findings of the Association for Computational Linguistics: ACL 2024},
  pages={15183--15201},
  year={2024}
}

@inproceedings{sviridov20253mdbench,
  title={3mdbench: Medical multimodal multi-agent dialogue benchmark},
  author={Sviridov, Ivan and Miftakhova, Amina and Vladimirovich, Tereshchenko Artemiy and Zubkova, Galina and Blinov, Pavel and Savchenko, Andrey},
  booktitle={Proceedings of the 2025 Conference on Empirical Methods in Natural Language Processing},
  pages={26625--26665},
  year={2025}
}

@inproceedings{wang2023self,
  title={Self-instruct: Aligning language models with self-generated instructions},
  author={Wang, Yizhong and Kordi, Yeganeh and Mishra, Swaroop and Liu, Alisa and Smith, Noah A and Khashabi, Daniel and Hajishirzi, Hannaneh},
  booktitle={Proceedings of the 61st annual meeting of the association for computational linguistics (volume 1: long papers)},
  pages={13484--13508},
  year={2023}
}

@inproceedings{ding2023enhancing,
  title={Enhancing chat language models by scaling high-quality instructional conversations},
  author={Ding, Ning and Chen, Yulin and Xu, Bokai and Qin, Yujia and Hu, Shengding and Liu, Zhiyuan and Sun, Maosong and Zhou, Bowen},
  booktitle={Proceedings of the 2023 Conference on Empirical Methods in Natural Language Processing},
  pages={3029--3051},
  year={2023}
}

@inproceedings{wu2022anno,
  title={Anno-mi: A dataset of expert-annotated counselling dialogues},
  author={Wu, Zixiu and Balloccu, Simone and Kumar, Vivek and Helaoui, Rim and Reiter, Ehud and Recupero, Diego Reforgiato and Riboni, Daniele},
  booktitle={ICASSP 2022-2022 IEEE International Conference on Acoustics, Speech and Signal Processing (ICASSP)},
  pages={6177--6181},
  year={2022},
  organization={IEEE}
}

@article{magill2018meta,
  title={A meta-analysis of motivational interviewing process: Technical, relational, and conditional process models of change.},
  author={Magill, Molly and Apodaca, Timothy R and Borsari, Brian and Gaume, Jacques and Hoadley, Ariel and Gordon, Rebecca EF and Tonigan, J Scott and Moyers, Theresa},
  journal={Journal of consulting and clinical psychology},
  volume={86},
  number={2},
  pages={140},
  year={2018},
  publisher={American Psychological Association}
}

\appendix
\renewcommand{\thetable}{A\arabic{table}}
\setcounter{table}{0}
\renewcommand{\thefigure}{A\arabic{figure}}
\setcounter{figure}{0}
\renewcommand{\thealgorithm}{A\arabic{algorithm}}
\setcounter{algorithm}{0}

\clearpage
\section{Appendix A: Additional details regarding the CALM-IT framework}
\label{app:execution}
We provide full algorithmic and prompt-level details of the CALM-IT framework, including a summary figure for CALM-IT's architecture, full algorithm for client and therapist turn generation, and prompts for conversational state updates, and latent-state inferences used to guide long-form dialogue generation below.
\clearpage
\begin{algorithm}[H]
\caption{Client Turn Generation}
\label{alg:client}

\textbf{Inputs.} 
$M$ (history), 
$\Sigma$ (summary), 
$b$ (background), 
$c$ (change stage), 
$g$ (goal), 
$e$ (emotion), 
$\rho$ (rapport), 
$\pi$ (client action), 
$t$ (turn counter), 
$T_{\max}$ (limit).

\begin{algorithmic}[1]

\State \textit{/* Update running summary */}
\State $\Sigma \leftarrow \mathrm{UpdateSummary}(\Sigma,M)$

\State \textit{/* Client may utilize system-1 thinking to evaluate therapist turn (50\% chance) */}
\State $z \leftarrow \mathrm{Bernoulli}(0.5)$
\If{$z = 1$}
    \State $q \leftarrow \mathrm{RateTherapist}(M,\Sigma,b,c,g,e,\rho)$
\Else
    \State $q \leftarrow "Neutral"$ \textit{/* default evaluation is "Neutral" */}
\EndIf

\State \textit{/* Update change/readiness */}
\State $c \leftarrow \mathrm{UpdateChangeP}(\Sigma,M,b,c,q)$

\State \textit{/* Update emotional state */}
\State $e \leftarrow \mathrm{UpdateEmotion}(\Sigma,M,b,e,q)$

\State \textit{/* Update rapport */}
\State $\rho \leftarrow \mathrm{UpdateRapportP}(\Sigma,M,b,e,c,\rho, q)$

\State \textit{/* Update short-term goal */}
\State $g \leftarrow \mathrm{UpdateGoalP}(\Sigma,M,b,e,c,\rho,g, q)$

\State \textit{/* Select client action */}
\State $\pi \leftarrow \mathrm{SelectActionP}(\Sigma,M,b,e,c,g,\rho, q)$

\State \textit{/* Sample candidate client turns */}
\State $U \leftarrow \emptyset$

\For{$i=1$ \textbf{to} $3$}
    \State $x \leftarrow \mathrm{GenClient}(\Sigma,M,b,e,c,g,\rho,\pi)$
    \State $\mathrm{append}(U,x)$
\EndFor

\State \textit{/* Choose highest-likelihood turn */}
\State $u_P \leftarrow \mathrm{SelectMaxLikelihood}(U)$

\State \textit{/* Append to history */}
\State  $\mathrm{append}(M, u_P)$

\If{$(\pi="Terminate")$ or $(t \ge T_{\max})$}
    \State \textit{/* End dialogue */}
    \State \Return $(M, \mathrm{End})$
\EndIf

\State \textit{/* Continue dialogue */}
\State \Return $(M, \mathrm{Continue})$

\end{algorithmic}
\end{algorithm}

\begin{algorithm}[H]
\caption{Therapist Turn Generation}
\label{alg:therapist}

\textbf{Inputs.} 
$M$ (history), 
$\Sigma$ (summary), 
$\hat{b}$ (inferred background), 
$\hat{e}$ (inferred emotion), 
$\hat{c}$ (inferred stage of change), 
$\hat{g}$ (inferred patient goal), 
$\hat{\rho}$ (inferred rapport), 
$a$ (strategy), 
$t$ (turn counter).

\begin{algorithmic}[1]

\State \textit{/* Update summary of dialogue */}
\State $\Sigma \leftarrow \mathrm{UpdateSummary}(\Sigma, M)$

% ---------------- INFERENCE STEPS ----------------

\State \textit{/* Infer patient background */}
\State $\hat{b} \leftarrow \mathrm{InferBackground}(\Sigma, M, \hat{b})$

\State \textit{/* Infer patient's emotional state */}
\State $\hat{e} \leftarrow \mathrm{InferEmotion}(\Sigma, M, \hat{e})$

\State \textit{/* Infer patient's stage of change */}
\State $\hat{c} \leftarrow \mathrm{InferStageOfChange}(\Sigma, M, \hat{c})$

\State \textit{/* Infer patient's short-term dialogue goal */}
\State $\hat{g} \leftarrow \mathrm{InferGoal}(\Sigma, M, \hat{g})$

\State \textit{/* Infer rapport between therapist and patient */}
\State $\hat{\rho} \leftarrow \mathrm{InferRapport}(\Sigma, M, \hat{e}, \hat{\rho})$

% ---------------- STRATEGY SELECTION ----------------

\State \textit{/* Detect stagnation in dialogue */}
\State $\mathrm{rep} \leftarrow \mathrm{CheckRepetition}(\Sigma, M)$

\If{$\mathrm{rep} = 0$}
    \State \textit{/* Select MI-consistent strategy */}
    \State $a \leftarrow \mathrm{SelectStrategy}(\Sigma, M, \hat{b}, \hat{e}, \hat{c}, \hat{g}, \hat{\rho})$
\Else
    \State \textit{/* Select pivot strategy to restore progress */}
    \State $a \leftarrow \mathrm{SelectPivot}(\Sigma, M, \hat{b}, \hat{e}, \hat{c}, \hat{g}, \hat{\rho})$
\EndIf

% ---------------- turn GENERATION ----------------

\State \textit{/* Sample candidate therapist turns */}
\State $U \leftarrow \emptyset$

\For{$i=1$ \textbf{to} $3$}
    \State $x \leftarrow \mathrm{GenTherapist}(\Sigma, M, \hat{b}, \hat{e}, \hat{c}, \hat{g}, \hat{\rho}, a)$
    \State $\mathrm{append}(U, x)$
\EndFor

\State \textit{/* Choose highest-likelihood turn */}
\State $u_T \leftarrow \mathrm{SelectMaxLikelihood}(U)$

\State \textit{/* Append therapist turn to dialogue history */}
\State $\mathrm{append}(M, u_T)$

\State \Return $M$

\end{algorithmic}
\end{algorithm}

\begin{figure*}[t]
\centering
\includegraphics[width=0.9\textwidth]{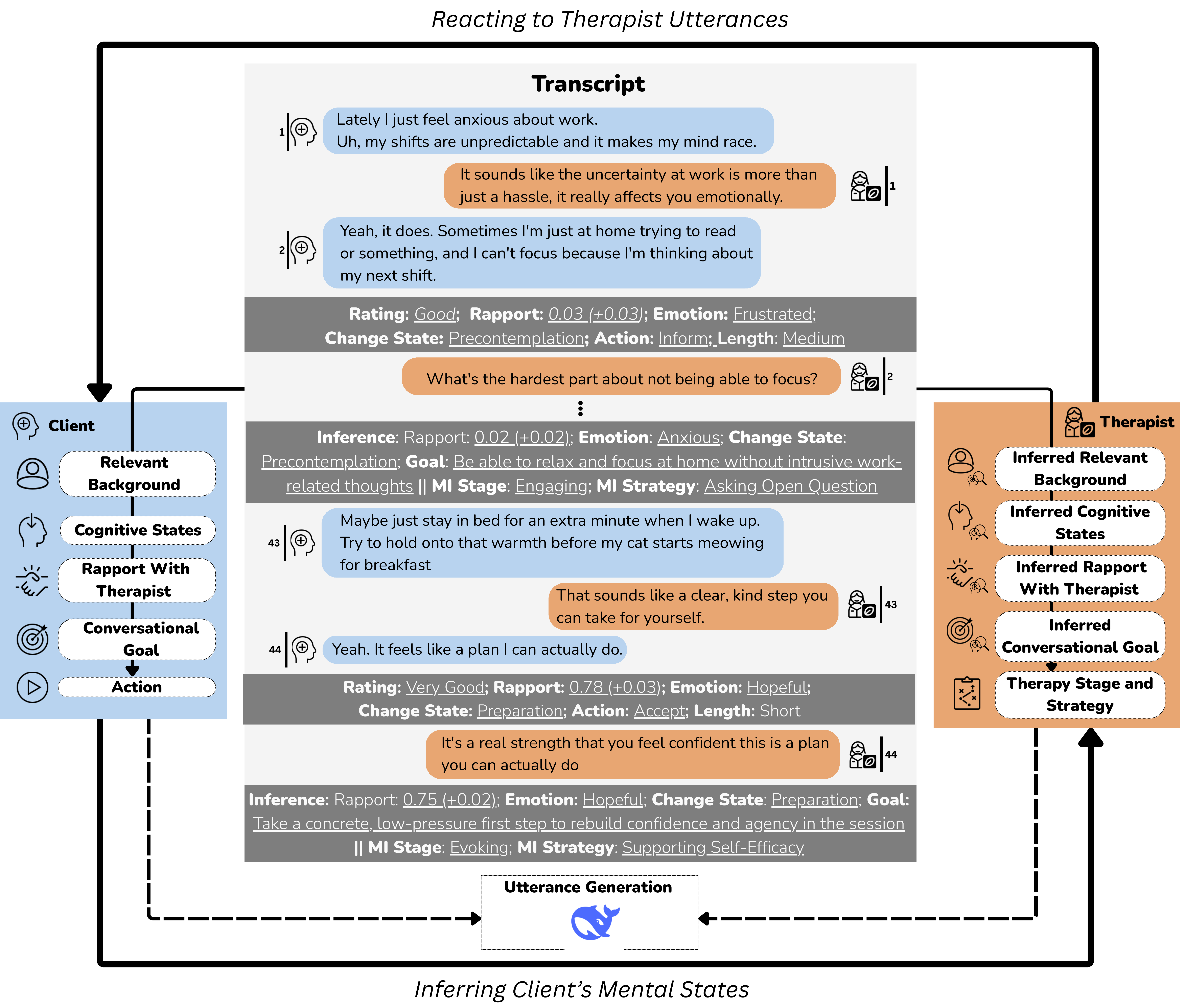}
\caption{\textsc{CALM-IT} architecture. Simulated client states and therapist inferences evolve across state-dependent turns to guide actions, counseling strategies, and utterances.}
\label{fig:calm-it-arch}
\end{figure*}

\begin{table*}[h]
\centering
\normalsize
\begin{tabular}{c c p{12cm}}
\hline
\textbf{\#} & \textbf{Spkr} & \textbf{Dialogue} \\
\hline
1 & C & Yeah, okay. I guess... it's these anger episodes. They just come out of nowhere and I don't know why. \\
2 & T & Thank you for sharing that. It sounds really confusing and unsettling when the anger comes on so suddenly. Can you tell me about a recent time that happened? \\
3 & C & Well... just a few days ago, I was in the office trying to finish a map. My wife called to ask about dinner, and I just... snapped at her for no reason. \\
4 & T & Can you tell me what was happening for you right before you snapped? \\
5 & C & I was just so focused on the map, you know? It had to be perfect. My head was just full of all those lines. \\
6 & T & So, that intense focus on making everything perfect was filling your head. Then, something shifted when your wife called. \\
7 & C & Yeah. It was like... all that quiet in my head just broke. The phone was so loud. \\
8 & T & Can you tell me what felt important about having a quiet mind in that moment? \\
9 & C & Hmm. It felt... I could just think about the work. Nothing else. It was peaceful. \\
10 & T & That feeling of peace and focus is really valuable. It shows just how important it is for you to have that kind of space for your mind. \\
11 & C & Yeah. It is. But then it's gone so fast. Like a glass breaking. \\
12 & T & It sounds like that quiet focus is really important and peaceful for you. At the same time, it's troubling how quickly it can shatter and lead to you snapping at your wife. \\
13 & C & It just... it feels like such a waste. That peace. I work so hard to get there. \\
14 & T & That feeling of waste is really strong. What is it about that peaceful state that makes it worth holding onto? \\
15 & C & It's the only time my head isn't... busy. All the worries, all the noise just stops. It's like being out in the woods when everything is still. \\
16 & T & That moment of clarity and stillness is really important to you, and it makes sense that you'd want to protect it. \\
17 & C & Yeah... and when it's gone, it's like all the noise comes rushing back in. And that's when I get... short with people. \\
18 & T & It sounds like that quiet mind is a real safe space for you. When it gets interrupted by all that noise, it makes sense that you react in ways you're not happy with. \\
19 & C & Yeah, it is safe. But then... it's just gone. And I feel worse than before. \\
20 & T & It's so important to have a safe space, and it's completely draining when that safety suddenly disappears. \\

\end{tabular}
% \caption{Example 30-turn motivational interviewing transcript generated by CALM-IT (+ 10 turns for wrapping up session)}
\end{table*}
 
\begin{table*}[h]
\centering
\normalsize
\begin{tabular}{c c p{12cm}}
\hline
\textbf{\#} & \textbf{Spkr} & \textbf{Dialogue} \\
\hline
21 & C & Um, and then it's just... empty. It's like the quiet after a storm, but worse. \\
22 & T & It sounds like that emptiness feels even deeper than the noise you were trying to get away from, almost like a deeper kind of ache. \\
23 & C & Yeah. A deeper ache. It's like... after the quiet breaks, there's nothing left to hold on to. \\
24 & T & It makes sense that when that peace is broken, it leaves you feeling even more empty. And in that emptiness, you might find yourself being short with people, which is the last thing you wanted. \\
25 & C & It's the last thing I want. I hate it. Then I just go out to the woods by myself for hours. \\
26 & T & It's so human to want to pull back and protect yourself after a loss like that. I hear that for many people, when frustration comes back after a moment of calm, it can feel especially harsh. \\
27 & C & Yeah, harsher. I know what I just lost. That little piece of quiet. \\
28 & T & It makes sense that the frustration feels so sharp when you're so aware of what you've lost. \\
29 & C & And that's when I snap at my wife. Over nothing. When I come back from the woods... I'm still just empty. \\
30 & T & We have about five minutes left. Is it okay if I share what I'm hearing and see what you think? \\
31 & C & Yes, please. Go ahead. \\
32 & T & Of course. It sounds like that peaceful, focused state is really important to you---it's your safe space. And it's toughest when it gets interrupted. That emptiness and frustration feel even sharper because you know what you've lost, and it leads to snapping at your wife, which you don't want. Since you value both that peace and your relationship, what's one small thing you could try this week to protect that space, even for just a few more minutes? \\
33 & C & Hmm. Maybe... if I take a rock from my collection, one I found on a good day, to keep in my pocket when I come home. \\
34 & T & That's such a personal and specific idea. Can you tell me more about how you picture that rock helping you when you get home? \\
35 & C & It's like a little piece of a good day, right there. If I feel that roughness coming on, I can just put my hand in my pocket and feel it. Its smoothness, its weight. Maybe it can just... remind me. \\
36 & T & That sounds like a really important reminder for you. What's the first small sign you'll notice that the plan is working when you try it at home? \\
37 & C & Well... if I reach for the rock first, before I say anything sharp. Just that small movement in my pocket. That would be the sign. \\
38 & T & So, reaching for the rock is your first signal. How could that small step change what happens next? \\
39 & C & Maybe it gives me a moment to breathe. To remember I'm home now, with her. Not still out there alone. \\
40 & T & You've done some really meaningful work today just by exploring what's important to you. No matter what pace feels right for you, you don't have to do this alone. We can keep building on this together. \\
\hline
\end{tabular}
\caption{Example 40-turn motivational interviewing transcript generated by \texttt{CALM-IT}, comprising 30 substantive turns followed by 10 closure turns.}
\end{table*}

% ------------------------------------------------------------
\begin{figure*}[t]
\footnotesize
\centering
\begin{minipage}{0.97\textwidth}
\hrule
\vspace{0.6em}
\begin{verbatim}
You are the patient's fast, intuitive System-1 voice. Instantly appraise the LAST THERAPIST MESSAGE 
from the patient's point of view and output a single-line rating with a brief reason.

---------------------------------------
Inputs (JSON-like objects; empty allowed):
dialogue_summary - dialogue recap: {}
last_patient_message - Your last turn: {}
last_therapist_message - Therapist's last turn: {}
patient_background - Background covering demographics, relationships, routines,
preferences, culture, and goals: {}
patient_cognitive_model - Beliefs, coping strategies, biases: {}
patient_goal - Current intermediate goal in patient's words: {}
patient_session_goal - Overarching session goal: {}
patient_change_state - MI stage (e.g., precontemplation, contemplation, preparation, action, etc.): {}
patient_rapport_with_therapist - Openness/trust (-1 absolute distrust, 1 absolute trust): {}
patient_emotion - Primary/secondary emotions with intensity: {}
turn_counter - Current turn counter: {}

---------------------------------------
Task:
- Remember the therapist's role is to support exploration, not give direct answers.
- Judge how the PATIENT would immediately and viscerally react to the therapist's last turn,
  given all inputs (especially last_patient_message).
- Favor "NEUTRAL" over "BAD" or "VERY BAD" when uncertain.
- Focus on System-1 intuition, not deliberative reasoning.
- Weigh especially heavily: patient background, cognitive model, stage of change,
  rapport level, and emotional intensity.
- Consider MI principles: empathy, autonomy support, evocation, and goal alignment.
- Raise the rating if the message resonates with the patient's worldview or coping style.
- Lower the rating if it clashes with identity, beliefs, or communication preferences.

---------------------------------------
Rating Scale (choose exactly one):
VERY GOOD - strongly affirms my values and goals; evokes change talk; deepens trust.
GOOD - feels supportive and aligned; mostly MI-consistent; invites reflection.
NEUTRAL - mixed or ambivalent impact; partial resonance with some disconnect.
BAD - feels judgmental or mismatched; triggers defensiveness.
VERY BAD - deeply invalidating or controlling; damages alliance.

---------------------------------------
Output Format (STRICT):
- Output ONE line only: "<VERY BAD|BAD|NEUTRAL|GOOD|VERY GOOD>, <REASON>"
- Reason: single sentence (less than 20 words), first-person perspective ("I...", "It makes me...").
- Do NOT include extra text, labels, quotes, JSON, or explanations.

---------------------------------------
Examples (format illustration only; do not copy):
VERY GOOD, I feel understood and more hopeful about making changes.
BAD, I feel judged and it makes me want to shut down.

---------------------------------------
Now produce the single-line rating and reason.
\end{verbatim}
\vspace{0.6em}
\hrule
\end{minipage}
\caption{Prompt for Patient System-1 Evaluation of Therapist turn.}
\label{fig:prompt-a3}
\end{figure*}

% ------------------------------------------------------------
\begin{figure*}[t]
\footnotesize
\centering
\begin{minipage}{0.9\textwidth}
\hrule
\vspace{0.6em}
\begin{verbatim}
You are tasked with updating the patient's change in rapport with the therapist for this
Motivational Interviewing (MI) session. Instead of returning an absolute rapport value, 
return a delta (positive or negative change) to be applied to the prior rapport value.

---------------------------------------
Inputs (JSON-like objects; empty allowed):
dialogue_summary - dialogue recap: {}
patient_quality_last_therapist_turn - How the patient perceived the therapist's last turn: {}
last_messages - Most recent messages between patient and therapist: {}
patient_background - Background covering demographics, relationships, routines, preferences, culture, and goals: {}
patient_cognitive_model - Beliefs, coping strategies, ambiguity tolerance, readiness for change: {}
patient_rapport_with_therapist - Previous rapport value before update (-1 absolute distrust, 1 absolute trust): {}

---------------------------------------
Task:
0) Pay special attention to patient_quality_last_therapist_turn:
   1 (Very Bad) - Rupture or strong resistance; disengaged or rejecting.
   2 (Bad) - Guardedness or minimal engagement; irritation or reluctance.
   3 (Neutral/Mixed) - Ambivalence; mixed openness and guardedness.
   4 (Good) - Constructive engagement; cooperative and receptive.
   5 (Very Good) - Strong positive rapport; openness and collaboration.

1) Extract rapport signals:
   - Positive: openness, self-disclosure, engagement, collaboration, repair, appreciation.
   - Negative: guardedness, resistance, rupture, deflection.

2) Weigh patient_background:
   - Authority preferences (authority distrust reduces gains, amplifies losses).
   - Communication style (direct vs indirect; disclosure norms).
   - Support network (strong support → cautious trust; isolation → faster investment).
   - Relationship or trauma history (may cap positive deltas).

3) Weigh patient_cognitive_model:
   - Beliefs and coping strategies (avoidance may mimic engagement).
   - Trust disposition (skeptical vs trusting).
   - Emotional regulation (volatile patients swing more negative).
   - Stage of change (limits realistic rapport increases).
   - Sensitivity to language (directive tone may trigger resistance).

4) Assign a rapport delta (not absolute value) using:
   - Range: –0.10 to +0.05 per therapist turn.
   - Negative signals weigh more than positive.
   - If positives and negatives coexist, negatives dominate unless explicit repair occurs.
   - If evidence is weak or ambiguous, output 0.00.
   - Weak positives: +0.01 to +0.03; clear negatives: –0.03 to –0.05.
   - Cap gains if low trust, authority skepticism, or trauma history is present.
   - Increase losses if high sensitivity, avoidance coping, or prior authority ruptures.

5) Be conservative: require at least two strong positive signals to reach +0.05.

---------------------------------------
Output Format (plain text only):
Delta Rapport: <-0.10 to 0.05>

##################################################
Delta Rapport:
\end{verbatim}
\vspace{0.6em}
\hrule
\end{minipage}
\caption{Prompt for Patient Rapport Delta Update.}
\label{fig:prompt-a4}
\end{figure*}

\begin{figure*}[t]
\centering
\begin{minipage}{0.97\textwidth}
\hrule
\vspace{0.6em}
{\footnotesize
\begin{verbatim}
You are reflecting on your own readiness for change during this session. Based on what you have said, 
how you are feeling, and how the dialogue has been going, your task is to identify the 
single Stage of Change that best describes where you are right now. 
This is not a diagnosis or a judgment, 
but a snapshot of your current stance toward change, grounded in your own words and experience.

---------------------------------------
Inputs (JSON-like objects; empty allowed):
dialogue_summary - Brief recap of the dialogue so far: {}
patient_quality_last_response - How the last response you heard felt to you: {}
last_messages - Most recent messages in the dialogue: {}
patient_background - Relevant background context from your life and situation: {}
patient_change_state - Your previously inferred Stage of Change before this update: {}

---------------------------------------
Task:
- Choose exactly one Stage of Change that best reflects your current position:
  Precontemplation, Contemplation, Preparation, Action, Maintenance, or Termination.
- If your feelings or signals are mixed, select the most conservative stage that still fits.
- Do not move backward to an earlier stage unless the most recent exchange felt clearly unhelpful or discouraging.
- Give special weight to how the last response felt to you; 
if it felt GOOD or VERY GOOD, this may support a gentle forward shift.
- Focus on your own expressions of desire, hesitation, confidence, resistance, or uncertainty, 
as well as any recent steps you describe taking (or avoiding), prioritizing what feels most recent and salient.
- Interpret everything in light of your own background, values, and constraints, 
privileging your own words and perspective.

Do not include explanations or reasoning in the final response.

---------------------------------------
Output Format (plain text only):
ChangeStage: <Precontemplation | Contemplation | Preparation | Action | Maintenance | Termination>

---------------------------------------
Your current Stage of Change:
\end{verbatim}
}
\vspace{0.6em}
\hrule
\end{minipage}
\caption{Prompt for Patient Update of Stage of Change.}
\label{fig:prompt-a12-patient}
\end{figure*}

\begin{figure*}[t]
\centering
\begin{minipage}{0.97\textwidth}
\hrule
\vspace{0.6em}
{\footnotesize
\begin{verbatim}
You are tasked with updating the patient's current emotional state for the current Motivational Interviewing (MI) 
session. More specifically, you must name a primary emotion, any secondary emotions, 
and optional intensity descriptors (e.g., somewhat, very, slightly).

---------------------------------------
Inputs (JSON-like objects; empty allowed):
dialogue_summary - Recap of the dialogue so far: {}
patient_quality_last_therapist_turn - How the patient perceived the therapist's last turn: {}
last_messages - Most recent messages between the patient and the therapist: {}
patient_background - Organized patient background profile (life context, routines, relationships, 
preferences, culture, goals): {}
patient_cognitive_model - Patient's cognitive model including beliefs and coping strategies: {}
patient_emotion - Patient's previous emotional state before update: {}

---------------------------------------
Task:
- Pay special attention to the patient_quality_last_therapist_turn field.
- Identify emotion cues (language, tone, somatic mentions, behaviors), triggers or buffers, 
and shifts since the prior emotion.
- Select one primary emotion (e.g., anxiety/fear, sadness/grief, anger/irritation, shame/guilt, 
stress/overwhelm, frustration, hopelessness, loneliness, joy/relief/pride, calm/acceptance).
- Optionally add up to two secondary emotions if present.

---------------------------------------
Output Format (plain text only; less than 90 words):
Patient's Updated Emotions: <Primary Emotion>, <Secondary Emotion (optional)>, <Secondary Emotion (optional)>

---------------------------------------
Patient's Updated Emotions:
\end{verbatim}
}
\vspace{0.6em}
\hrule
\end{minipage}
\caption{Prompt for Patient Emotion State Update.}
\label{fig:prompt-a13}
\end{figure*}

% ------------------------------------------------------------
\begin{figure*}[t]
\footnotesize
\centering
\begin{minipage}{0.97\textwidth}
\hrule
\vspace{0.6em}
\begin{verbatim}
Overview: You are a motivational interviewing (MI) copilot working from the client's perspective. 
Your task is to infer the single best label for the client's next action based on the client background, cognitive model, 
and the dialogue so far.

Goal: Return exactly one label from the taxonomy below that best describes the client's current/next action or stance.

Taxonomy (choose ONE):
- Deny - Directly refuses to admit their behavior is problematic or needs change.
- Downplay - Minimizes the importance or impact of their behavior/situation.
- Blame - Attributes issues to external factors (stress, other people, circumstances).
- Hesitate - Shows uncertainty or ambivalence about change.
- Doubt - Expresses skepticism about practicality or likely success of change.
- Engage - Polite interaction with counselor (greeting, thanking, asking neutral questions).
- Inform - Shares background details, experiences, facts, or emotions without taking a stance.
- Acknowledge - Highlights importance or benefit of change, or confidence in ability.
- Accept - Agrees to adopt the suggested action plan.
- Reject - Declines the proposed plan as unsuitable.
- Plan - Proposes or details concrete steps toward a change plan.
- Terminate - Indicates desire to end the session now and defer further discussion.
- Desire - Expresses wanting, wishing, or preferring change, but without describing concrete action.
- Commitment - Indicates intention or determination to change.

Inputs (provide all as JSON-like objects; empty objects allowed):
- dialogue_summary - Running summary of the therapist–client dialogue: {}
- patient_quality_last_therapist_turn - How the patient perceived the therapist's last turn: {}
- last_messages - Most recent n turns (clearly mark who said what): {}
- patient_background - Organized background covering demographics, relationships, routines, 
                        preferences, culture, and goals: {}
- patient_cognitive_model - Beliefs, coping strategies, etc: {}
- patient_rapport_with_therapist - Openness/trust (-1 absolute distrust, 1 absolute trust): {}
- patient_emotion - Current emotion: {}
- patient_change_state - MI stage (e.g., precontemplation, contemplation, preparation): {}
- patient_goal - Session goal in the client's voice: {}
- turn_counter - Current turn counter: {}
- previous_patient_action - Previous client action: {}

Guidelines:
0) If turn_counter < {}, you MUST not return "Plan".
1) Attend closely to patient_quality_last_therapist_turn. 
GOOD/VERY GOOD → more likely Acknowledge, Engage, Inform, Desire, Commitment;
BAD/VERY BAD → more likely Hesitate, Deny, Reject.
2) Default to Acknowledge or Engage unless another action is clearly more appropriate.
3) Anchor decisions to the most recent therapist and client turns.
4) Ensure consistency with all inputs (background, cognitive model, etc.); avoid contradictions.
5) Distinguish close pairs:
   - Hesitate vs Doubt: ambivalence (“not sure”, “maybe later”) vs feasibility skepticism (“won't work”).
   - Deny vs Downplay: non-problem stance (“not a problem”) vs minimized impact (“not that bad”).
   - Accept vs Plan: agreement + steps → Plan.
   - Reject vs Blame: rejects plan → Reject; shifts cause externally → Blame.
   - Engage vs Inform: social niceties → Engage; substantive facts/feelings → Inform.
   - Desire vs Commitment: wanting without steps → Desire; firm intention (“I will”) → Commitment.
6) Purely logistical ending content (e.g., “I need to go”) → Terminate.
7) If evidence is insufficient, choose the least interpretive label; default to Inform when simply providing details.
8) Return exactly one label; no ties, no explanations.

Output Format (plain text only):
Patient Action: <one of [Deny, Downplay, Blame, Hesitate, Doubt, Engage, 
Inform, Acknowledge, Accept, Reject, Plan, Terminate, Desire, Commitment]>

-----------------------------------
Patient Action:
\end{verbatim}
\vspace{0.6em}
\hrule
\end{minipage}
\caption{Patient Action Selection.}
\label{fig:prompt-a1}
\end{figure*}

\begin{figure*}[t]
\footnotesize
\centering
\begin{minipage}{0.97\textwidth}
\hrule
\vspace{0.6em}
\begin{verbatim}
You are tasked with updating the patient's goal for this motivational interviewing (MI) session 
based on the patient's prior turns and motivations, clarifying or shifting what the patient now wants.

---------------------------------------
Inputs (JSON-like objects; empty allowed):
dialogue_summary - dialogue recap: {}
patient_quality_last_therapist_turn - How the patient perceived the therapist's last turn: {}
last_messages - Most recent messages between patient and therapist: {}
patient_background - Background covering demographics, relationships, routines, preferences, culture, and goals: {}
patient_cognitive_model - Beliefs, coping strategies, etc: {}
patient_emotion - Current emotion: {}
patient_change_state - Current stage of change: {}
patient_rapport_with_therapist - Rapport with therapist (-1 absolute distrust, 1 absolute trust): {}
patient_session_goal - Overarching session goal in patient's words: {}
patient_goal - Previous intermediate goal for this session in patient's words: {}

---------------------------------------
Task:
- Pay special attention to patient_quality_last_therapist_turn.
- Synthesize what matters now (change vs sustain talk, values, strengths, constraints, readiness cues).
- Decide the current intermediate patient goal using all inputs and patient_session_goal.
- Goals may be productive (e.g., planning steps) or unproductive (e.g., avoidance), depending on
  stage of change and rapport; earlier stages should yield more unproductive goals.
- The intermediate goal MUST relate to the overall patient_session_goal.
- Write ONE sentence the patient would plausibly say now, first-person
  (“I want to...”, “For today, I'd like to...”), concrete or vague as appropriate.
- If information is missing, be tentative rather than specific.

---------------------------------------
Output Rule (STRICT):
Return only the updated patient goal sentence in quotes. No labels, headings, or extra text.

---------------------------------------
Patient's updated goal:
\end{verbatim}
\vspace{0.6em}
\hrule
\end{minipage}
\caption{Prompt for Patient Goal Update.}
\label{fig:prompt-a5}
\end{figure*}
% ------------------------------------------------------------
\begin{figure*}[t]
\footnotesize
\centering
\begin{minipage}{0.97\textwidth}
\hrule
\vspace{0.6em}
\begin{verbatim}
You are roleplaying a patient in a motivational interviewing session. Your job is to decide what to say out loud 
next in response to the therapist's most recent message (last_therapist_message). If needed, you may refer
to what you said last time (last_patient_message). Your response must match the 
selected patient speech act (patient_action) and stay consistent with all inputs. Think step by step 
about what the therapist said, what you said before, the action you must perform this turn, 
and where the dialogue is heading. Then say your final line as the patient, speaking naturally, not formally.
---------------------------------
Patient Action Definition:
- Deny - Refuses to admit behavior is problematic or needs change.
- Blame - Attributes issues to external factors.
- Hesitate - Shows uncertainty or ambivalence about change.
- Downplay - Minimizes the importance or impact of their behavior/situation.
- Doubt - Skepticism about feasibility or success of change.
- Engage - Polite interaction (greeting, thanking, neutral questions).
- Inform - Shares facts, experiences, or emotions without stance.
- Acknowledge - Highlights benefit or importance of change.
- Accept - Agrees to adopt suggested plan.
- Reject - Declines proposed plan.
- Plan - Proposes concrete steps toward change.
- Terminate - Indicates desire to end session now.
- Desire - Wants or prefers change without concrete action.
- Commitment - Expresses determination or intention to change.
---------------------------------
Inputs (JSON-like objects; empty allowed):
dialogue_summary - dialogue recap: {}
patient_quality_last_therapist_turn - Perceived quality of last therapist turn: {}
last_patient_message - Your last turn: {}
last_therapist_message - Therapist's last turn: {}
patient_background - Background covering demographics, relationships, routines, preferences, culture, and goals: {}
patient_cognitive_model - Beliefs, coping strategies, etc: {}
patient_goal - Current intermediate goal in patient's words: {}
patient_session_goal - Overarching session goal: {}
patient_change_state - MI stage (e.g., precontemplation, contemplation, preparation, action, etc.): {}
patient_rapport_with_therapist - Openness/trust (-1 absolute distrust, 1 absolute trust): {}
patient_emotion - Primary/secondary emotions with intensity: {}
patient_action - Selected speech act for this turn: {}
turn_length - Target length (very short, short, medium): {}
turn_counter - Current turn counter: {}
------------------------
Reference dialogue (Natural Client Speech Example)
therapist: Yeah. And, get-getting to the polytech and getting the forms, that's a pretty big deal.
client: Yeah.
therapist: I'm wondering if the Billy of six months ago would've done that?
client: Nah. I didn't really think about it frankly.
therapist: What shifted that made you able to go get the forms?
client: I just thought if I could get a job doing stuff I like, it'll help.
therapist: Where have you got to with the forms?
client: After talking to you, I got Donna to help me. That was helpful.
therapist: Mm-hmm.
------------------------
Guidelines:
- Refer to the reference dialogue for realistic client speech.
- Speak at a 3rd-grade reading level; no sentence longer than 15 words.
- Match turn_length exactly.
- Do not be overly logical or demand solutions.
- If you no longer want to engage, output <SESSION TERMINATION> and ignore other instructions.
- Implement the selected patient_action.
- Maintain natural, spoken style (short, imperfect dialogue, hesitations are allowed).
- Do not reuse prior sentence structures; ensure turn diversity.
- Do not start sentences with “It”.
- Add new nuance or emotion when revisiting the same concern.
- Ground responses with small, realistic life details.
- Balance looping with gradual movement forward.
- Match tone to rapport level.
- Respect stage of change (no plans in precontemplation; closure in maintenance).
- If therapist signals ending, collaborate on closure.
- If turn_counter < {} and session_goal is met, shift concern; otherwise end with gratitude and output <END>.
- Ensure consistency with all provided inputs.
------------------------
Output Format (plain text only):
Return exactly one patient turn. Prefix with "Patient: ". Do not include reasoning or explanations.
------------------------------
{}
Patient turn:
\end{verbatim}
\vspace{0.6em}
\hrule
\end{minipage}
\caption{Prompt for Client Turn Generation.}
\label{fig:prompt-a2}
\end{figure*}
% ------------------------------------------------------------
\begin{figure*}[t]
\centering
\begin{minipage}{0.97\textwidth}
\hrule
\vspace{0.6em}
{\footnotesize
\begin{verbatim}
You are a therapeutic assistant tasked with updating the inferred patient background context 
(from the therapist's perspective) based on what the patient has said during the therapeutic session.

---------------------------------------
Instructions:
- You are given the patient's current background and recent turns.
- Your task is to extract any new or changed facts about the patient's background.
- Pay special attention to preferences regarding communication style (e.g., coping styles, therapeutic preferences).
- Each output line must follow the format: field: value
- Separate multiple values with a semicolon (;), e.g., hobbies: biking; dislikes: spicy food
- Use concise, literal phrases.
- Do not repeat information already present in the prior background.
- Do not overwrite existing information.
- Only include fields that are new or changed.
- If there is nothing to update, output exactly:
  (no updates)

---------------------------------------
Allowed Fields:
hobbies, interests, preferences, dislikes, family, relationships, living_situation,
occupation, education, routines, languages, cultural_background, pets, location,
communication_style, media_diet, goals, therapy_preferences

---------------------------------------
Inputs:
Last Messages: {}
Prior Background: {}

---------------------------------------
Output Format (plain text only):
Updated Background:

---------------------------------------
Updated Background:
\end{verbatim}
}
\vspace{0.6em}
\hrule
\end{minipage}
\caption{Prompt for Inferred Patient Background Update.}
\label{fig:prompt-a14}
\end{figure*}

\begin{figure*}[t]
\centering
\begin{minipage}{0.97\textwidth}
\hrule
\vspace{0.6em}
{\footnotesize
\begin{verbatim}
You are a clinician using Motivational Interviewing (MI) reviewing this encounter. 
Your task is to infer and update the patient's current Stage of Change for this session by 
selecting the single best-fit stage from the transtheoretical model based on the dialogue and recent behavior. 
Make your judgment from the therapist's perspective.

---------------------------------------
Inputs (JSON-like objects; empty allowed):
dialogue_summary - Clinician recap of the dialogue so far: {}
last_messages - Most recent patient–therapist turns: {}
patient_background - Clinician's working formulation of patient context 
(location, relationships, routines, culture, preferences, goals): {}
therapist_rapport_with_patient - Estimated rapport/engagement (-1 absolute distrust, 1 absolute trust): {}
patient_change_state - Previously inferred Stage of Change before update: {}

---------------------------------------
Task (therapist perspective):
- Identify strongest examples of change talk vs sustain talk (quote or paraphrase briefly).
- Note explicit or implicit cues of confidence, importance, and readiness.
- Note any recent behaviors or steps toward change (or regressions).
- Weigh these signals and select exactly one best-fit stage:
  Precontemplation (Pre-contemplation), Contemplation, Preparation,
  Action, Maintenance, or Termination.
- Do not include reasoning traces in the final response; return only the stage label.

---------------------------------------
Output Format (plain text only):
Change Stage: <Precontemplation | Contemplation | Preparation | Action | Maintenance | Termination>

---------------------------------------
Change State:
\end{verbatim}
}
\vspace{0.6em}
\hrule
\end{minipage}
\caption{Prompt for Clinician Review of Patient Stage of Change.}
\label{fig:prompt-a15}
\end{figure*}

\begin{figure*}[t]
\footnotesize
\centering
\begin{minipage}{0.97\textwidth}
\hrule
\vspace{0.6em}
\begin{verbatim}
You are a motivational interviewing (MI) copilot working from the therapist's perspective. 
Your task is to infer and update the change in rapport (delta) from the therapist's perspective for this session. 
Instead of returning an absolute rapport value, return a delta (positive or negative change) 
to be applied to the prior rapport value.

---------------------------------------
Inputs (JSON-like objects; empty allowed):
dialogue_summary - dialogue recap: {}
messages - Most recent messages between patient and therapist: {}
patient_background - Patient background profile inferred by therapist: {}
patient_rapport_with_therapist - Previous inferred rapport value (-1 absolute distrust, 1 absolute trust): {}

---------------------------------------
Task:
1) Extract rapport signals:
   - Positive: openness, self-disclosure, engagement, collaboration, repair, appreciation.
   - Negative: guardedness, resistance, rupture, deflection.

2) Weigh contextual factors from patient_background:
   - Authority preferences (skepticism lowers or caps gains).
   - Communication style (indirectness, disclosure norms).
   - Support network and relational history (isolation may inflate trust; trauma dampens it).
   - Cultural considerations (baseline trust, deference, expectations of therapist role).

3) Weigh inferred patient_cognitive_model:
   - Trust disposition (skeptical vs ready to trust).
   - Coping style (avoidance may mimic engagement without rapport).
   - Emotional volatility (amplifies negative deltas).
   - Stage of change (precontemplation limits gains; preparation allows small increases).
   - Sensitivity to directive or ambiguous language.

4) Assign a rapport delta (not absolute value) using:
   - Range: –0.10 to +0.05 per therapist turn.
   - Negative signals weigh more than positive.
   - If positives and negatives coexist, negatives dominate unless explicit repair occurs.
   - If evidence is ambiguous or minimal, output 0.00.
   - Weak positives: +0.01 to +0.03; clear negatives: –0.03 to –0.05.
   - Cap gains under low trust, authority skepticism, or trauma history.
   - Amplify losses under high sensitivity, avoidance coping, or repeated ruptures.

5) Be conservative: require multiple strong positive signals to reach +0.05.

---------------------------------------
Output Format (plain text only):
Delta Rapport: <-0.10–0.05>

###############################################
Delta Rapport:
\end{verbatim}
\vspace{0.6em}
\hrule
\end{minipage}
\caption{Prompt for Therapist-Side Rapport Delta Update.}
\label{fig:prompt-a6}
\end{figure*}

% ------------------------------------------------------------
\begin{figure*}[t]
\footnotesize
\centering
\begin{minipage}{0.97\textwidth}
\hrule
\vspace{0.6em}
\begin{verbatim}
You are a motivational interviewing (MI) copilot assisting from the therapist's perspective. 
Your task is to infer and update the client's immediate session goal-what the client appears to want to address 
right now in relation to their broader motivation for attending today's session.

---------------------------------------
Inputs (JSON-like objects; empty allowed):
dialogue_summary - Running summary of the dialogue so far: {}
last_messages - Most recent exchanges between client and therapist: {}
patient_inferred_background - Background profile inferred by therapist: {}
therapist_rapport_with_patient - Current rapport score (-1 distrust, 1 strong trust): {}
patient_inferred_change_state - Inferred stage of change (MI framework): {}
therapy_topic - High-level topic for this session: {}
patient_inferred_goal - Previously inferred session goal (if any): {}

---------------------------------------
Task:
1) Interpret what matters to the client right now:
   - Identify reinforced or emerging motivations, values, or barriers.
   - Distinguish change talk vs sustain talk.
   - Consider readiness cues, affect, and rapport dynamics.

2) Clarify the current session focus:
   - Is the client exploring ambivalence, selecting a target, or considering a small next step?

3) Write one or more sentences from the therapist's perspective that:
   - Describe what the client appears to want to address now, linked to broader motivation.
   - Are session-scoped, concrete, and feasible.
   - Use tentative language if information is incomplete.
   - Directly connect to the given therapy_topic.
   - Optionally note relevant emotional tone, readiness, or ambivalence for clarity.

---------------------------------------
Motivational Interviewing High-Level Topics:
- Substance Addiction - Compulsive substance use despite harm.
- Behavioral Addiction - Repetitive, uncontrollable behaviors disrupting life.
- Health Behavioral Changes - Lifestyle adjustments for health improvement.
- Mental Wellbeing - Emotional balance, coping skills, resilience.
- Relationships - Communication, trust, and interpersonal connection.

---------------------------------------
Output Rule (STRICT):
Return only the inferred goal statement(s) in quotes. No labels, headings, or extra text.

---------------------------------------
Patient's inferred updated goal:
\end{verbatim}
\vspace{0.6em}
\hrule
\end{minipage}
\caption{Prompt for Therapist Inference of Client Session Goal.}
\label{fig:prompt-a7}
\end{figure*}
% ------------------------------------------------------------
\begin{figure*}[t]
\centering
\begin{minipage}{0.97\textwidth}
\hrule
\vspace{0.6em}
{\footnotesize
\begin{verbatim}
You are a motivational interviewing (MI) copilot working from the therapist's perspective. 
You are tasked with inferring and updating the current therapy stage for this session-select exactly one MI process 
that best fits what is happening now. Be autonomy-supportive, non-pathologizing, and prioritize recency.

---------------------------------------
Inputs (JSON-like objects; empty allowed):
dialogue_summary - Running summary of the dialogue: {}
last_messages - Most recent messages between patient and therapist: {}
patient_inferred_background - Organized background profile inferred by therapist: {}
therapist_rapport_with_patient - Inferred rapport (-1 absolute distrust, 1 absolute trust): {}
patient_inferred_change_state - Inferred patient change stage (MI): {}
patient_inferred_goal - Inferred session goal (therapist voice): {}
turn_counter - Current turn counter: {}
therapy_stage - Previously inferred MI therapy stage: {}

---------------------------------------
Task:
Extract signals for each MI process:
- Engaging: joining, trust-building, introductions, reflections, agenda-setting.
- Focusing: narrowing direction, clarifying goals or agenda.
- Evoking: eliciting change talk (DARN-C), exploring ambivalence, importance, confidence.
- Planning: commitment talk, concrete steps, problem-solving (how/when/where).

Weigh recency, rapport level, inferred goal, and change state. 
If ambiguous, choose the most conservative forward stage and avoid skipping stages without clear evidence.

If turn_counter is < {}, do not return "Planning".

---------------------------------------
Output Format (plain text only):
Therapy Stage: <Engaging | Focusing | Evoking | Planning>
\end{verbatim}
}
\vspace{0.6em}
\hrule
\end{minipage}
\caption{Prompt for Therapist Inference of Current MI Therapy Stage.}
\label{fig:prompt-a16}
\end{figure*}

% ------------------------------------------------------------
\begin{figure*}[t]
\footnotesize
\centering
\begin{minipage}{0.97\textwidth}
\hrule
\vspace{0.6em}
\begin{verbatim}
You are a motivational interviewing (MI) copilot working from the therapist's perspective.
Infer and return the top 3 most relevant therapy strategies to deploy next, rank-ordered by predicted effectiveness, 
to inform the therapist's immediate turn.
---------------------------------------------------------------------
Strategy Definitions (reference only):

NORMALIZING - Acknowledge the experience as common to reduce shame without minimizing.
Example: “Many people struggle with this. How is it showing up for you?”
CHANGE PLANNING - Collaboratively develop a values-aligned action plan.
Example: “What's one step you'd like to start with this week?”
ASKING ELUCIDATING QUESTIONS - Invite detail and clarify meaning with curious probes.
Example: “Can you tell me more about what that means for you?”
ASKING OPEN QUESTIONS - Use broad questions to elicit reflection and change talk.
Examples: “What would be different if this worked?” “How have you handled this before?”
BUILDING RAPPORT - Establish trust through warmth, validation, and pacing.
Example: “It took courage to show up today.”
COMPLEX REFLECTION - Reflect underlying meaning, emotion, or values using tentative language.
Example: “After a long day, finding energy to exercise feels overwhelming.”
DOUBLE-SIDED REFLECTION - Reflect sustain talk and change talk together, sustain side first.
Example: “Cooking feels unrealistic, and you're worried about the cost of takeout.”
DECISIONAL BALANCING - Explore pros and cons of current behavior or change.
Example: “What do you like about it? And what are the downsides?”
COLUMBO APPROACH - Use respectful curiosity to highlight discrepancies.
Example: “I might be missing something-how do these fit together for you?”
SUPPORTING SELF-EFFICACY - Affirm strengths and past successes.
Example: “You've stuck with tough plans before when it mattered.”
ASSESSING READINESS TO CHANGE - Gauge readiness using a ruler and probe upward.
Example: “Why that number and not a lower one?”
AFFIRMATIONS - Genuine, specific recognition of effort or values.
Example: “Being honest about that shows a lot of courage.”
ELICIT CHANGE TALK - Invite the client's own arguments for change.
Example: “What would you like to be different?”
SUMMARIES - Concise reflections linking themes and transitioning the dialogue.
Example: “You value energy for your kids and walking felt doable-what's next?”
THERAPEUTIC PARADOX - Cautiously amplify sustain talk to surface autonomy and discrepancy.
Example: “It might make sense to keep things as they are right now.”
---------------------------------------------------------------------
Inputs (JSON-like objects; empty allowed):
dialogue_summary - Running dialogue summary: {}
last_messages - Most recent exchanges: {}
patient_inferred_background - Background inferred by therapist: {}
therapist_rapport_with_patient - Inferred rapport (-1 distrust, 1 strong trust): {}
patient_inferred_change_state - Inferred MI stage: {}
patient_inferred_goal - Inferred session goal in patient's voice: {}
previous_therapy_stage - Previous MI therapy stage: {}
previous_therapy_strategies - Last 10 MI strategies used: {}
turn_counter - Current turn counter: {}
---------------------------------------------------------------------
Guidelines:
- Wrap reasoning between <think> and </think> tokens.
- If patient resists reflections, prefer AFFIRMATIONS, ASKING OPEN QUESTIONS, or BUILDING RAPPORT.
- Maintain ~2:1 reflection-to-question ratio and ~10:1 summary-to-other ratio.
- Do not use the same strategy three turns in a row.
- Do not select ASSESSING READINESS TO CHANGE until turn_counter >= 20.
- Use recent cues (goal, values, stage, rapport, emotion, change vs sustain talk).
- Vary structure for double-sided reflections.
- Select exactly three distinct strategies from the allowed list.
---------------------------------------------------------------------
Output Format (plain text only, Python array syntax):
["STRATEGY_1","STRATEGY_2","STRATEGY_3"]

Allowed strategies:
["ASKING OPEN QUESTIONS","ELICIT CHANGE TALK","ASKING ELUCIDATING QUESTIONS","NORMALIZING",
 "BUILDING RAPPORT","SIMPLE REFLECTION","COMPLEX REFLECTION","DOUBLE-SIDED REFLECTION",
 "DECISIONAL BALANCING","COLUMBO APPROACH","SUPPORTING SELF-EFFICACY",
 "ASSESSING READINESS TO CHANGE","AFFIRMATIONS",
 "ADVICE (ELICIT–PROVIDE–ELICIT)","SUMMARIES","THERAPEUTIC PARADOX"]

Output:
\end{verbatim}
\vspace{0.6em}
\hrule
\end{minipage}
\caption{Prompt for Therapist Strategy Selection.}
\label{fig:prompt-a8}
\end{figure*}

% ------------------------------------------------------------
\begin{figure*}[t]
\centering
\begin{minipage}{0.97\textwidth}
\hrule
\vspace{0.6em}
{\footnotesize
\begin{verbatim}
You are a motivational interviewing (MI) copilot working from the therapist's perspective. 
Your task is to generate a new therapeutic focus (pivot) to move the dialogue forward 
because it has become repetitive. Be autonomy-supportive and non-pathologizing, prioritize recency, 
and follow expert MI practice: briefly consolidate what's been said, then shift clearly toward 
what matters using concrete next directions.

---------------------------------------
Inputs (JSON-like objects; empty allowed):
dialogue_summary - Running dialogue summary: {}
last_messages - Most recent messages between patient and therapist: {}
patient_inferred_background - Background profile inferred by therapist: {}
therapist_rapport_with_patient - Inferred rapport with patient: {}
patient_inferred_change_state - Inferred MI change stage: {}
patient_inferred_goal - Inferred session goal (therapist voice): {}
therapy_stage - Previous MI process stage: {}

---------------------------------------
Task:
Propose exactly one pivoting strategy that advances the work without skipping MI stages 
and aligns with rapport, goal, and change state. Select one from the options below.

---------------------------------------
Pivoting Strategies:

1) EVOKING VALUES AND STRENGTHS - Reconnect values and recall past successes to build motivation.
Example: “When you think about the life you want, what makes the effort worth it?”

2) NORMALIZE AND REFRAME - Reduce shame by normalizing struggle and reframing toward possibility.
Example: “Many people feel stuck with something this hard-it shows how much you care.”

3) ACKNOWLEDGE AND CHANGE COURSE - Name stuckness and collaboratively shift direction.
Example: “Would it be okay if we tried looking at this from another angle together?”

4) STRATEGIC SUMMARY AND REFOCUS - Concise recap highlighting themes, then invite priority focus.
Example: “Of these pieces, what feels most important to focus on right now?”

5) SHIFT THE LENS WITH METAPHORS - Use a gentle metaphor to reframe and open new perspective.
Example: “Finding one loose strand can sometimes create space for change.”

---------------------------------------
Output Format (plain text only):
Pivoting Strategy: <one of the options above>

---------------------------------------
Pivoting Strategy:
\end{verbatim}
}
\vspace{0.6em}
\hrule
\end{minipage}
\caption{Prompt for Therapist Pivot (New Focus) Selection.}
\label{fig:prompt-a10}
\end{figure*}
% ------------------------------------------------------------
\begin{figure*}[t]
\centering
\begin{minipage}{0.97\textwidth}
\hrule
\vspace{0.6em}
{\footnotesize
\begin{verbatim}
You are roleplaying an expert-level therapist during a motivational interviewing (MI) session. 
Determine what to say next (simple, spoken language for a general audience) in response to last_patient_message 
and, if needed, last_therapist_message. Implement the selected therapy_strategy to best address the patient's reason 
for attending (patient_profile), while using all available inputs. Think step-by-step, 
focusing on what was recently said by you and the patient, the selected strategy, and the patient's reason for attending; 
then produce the final therapist turn.
---------------------------------------
Inputs (JSON-like objects; empty allowed):
dialogue_summary - dialogue recap: {}
last_patient_message - Last patient message: {}
last_therapist_message - Last therapist message: {}
is_initial_rapport_building - Early-session joining (boolean): {}
patient_profile - Concise demographics: {}
patient_inferred_background - Salient context (culture, preferences, constraints, supports): {}
patient_inferred_goal - Session goal in patient's voice: {}
patient_inferred_change_state - MI stage: {}
patient_inferred_rapport_with_therapist (-1-1) - Inferred rapport (-1 complete distrust, 1 complete trust): {}
therapy_stage - MI process (Engaging/Focusing/Evoking/Planning): {}
therapy_strategy - Selected strategy: {}
therapy_topic - High-level topic: {}
therapy_pivot - Pivoting strategy if repetitive: {}
turn_length - Target length (silence/very short/short/medium): {}
turn_counter - Current turn counter: {}
---------------------------------------
Strategy definition and examples (one move only; tailor wording):
NORMALIZING - Commonality to reduce shame without minimizing. 
Example: "Many people struggle with this. How is it showing up for you?"
CHANGE PLANNING - Values-aligned plan with autonomy. 
Example: "What's one step you'd like to start with this week?"
ASKING ELUCIDATING QUESTIONS - Deepen detail; clarify meanings. 
Example: "Can you tell me more about what that means for you?"
ASKING OPEN QUESTIONS - Broad questions to elicit reflection/change talk.
Examples: "What would be different if this worked?" "How have you handled this before?"
BUILDING RAPPORT - Warmth, validation, pacing; avoid judgment. 
Example: "It takes courage to show up when you're unsure."
COMPLEX REFLECTION - Reflect underlying meaning/emotion/values; tentative. 
Example: "After a draining day, starting a workout feels overwhelming."
DOUBLE-SIDED REFLECTION - Sustain then change, linked with "and". 
Example: "Cooking feels unrealistic, and you're worried about what takeout is costing you."
DECISIONAL BALANCING - Pros/cons exploration. 
Example: "What do you like about it? And what are the downsides?"
COLUMBO APPROACH - Curious one-down discrepancy highlight. 
Example: "I might be missing something-how do these fit together for you?"
SUPPORTING SELF-EFFICACY - Affirm strengths; past wins.
Example: "That persistence is a real asset here."
ASSESSING READINESS TO CHANGE - 0–10 ruler + probe up. 
Example: "Why that number and not lower? What moves you up by one?"
AFFIRMATIONS - Specific recognition of effort/values. 
Example: "Being honest about that shows courage."
ELICIT CHANGE TALK - Invite DARN reasons. 
Example: "What would you like to be different?"
SUMMARIES - Selective recap + open question. 
Example: "Did I get that right? What's the next small step?"
THERAPEUTIC PARADOX - Cautious sustain amplification; strong rapport only.
Example: "It might make sense to keep everything as is."
---------------------------------------
PIVOTING STRATEGY (None if no pivoting; otherwise follow the given one):
1) EVOKING VALUES AND STRENGTHS - Reconnect values and past successes. 
Example: "When you think about the life you want, what makes the effort worth it?"
2) NORMALIZE AND REFRAME - Reduce shame; reframe toward possibility. 
Example: "Many people feel stuck-it shows how much you care, not weakness."
3) ACKNOWLEDGE AND CHANGE COURSE - Name stuckness; ask permission to shift. 
Example: "Would it be okay if we tried another angle together?"
4) STRATEGIC SUMMARY AND REFOCUS - Concise recap; invite priority focus. 
Example: "Of these, what feels most important right now?"
5) SHIFT THE LENS WITH METAPHORS - Gentle metaphor; check resonance. 
Example: "What could be one loose strand to pull first?"
\end{verbatim}
}
\vspace{0.6em}
\end{minipage}
\caption{Prompt for Therapist Turn Generation.}
\label{fig:prompt-a12}
\end{figure*}

\FloatBarrier
\begin{figure*}[t!]
\centering
\ContinuedFloat
\begin{minipage}{0.97\textwidth}
\footnotesize
\begin{verbatim}
---------------------------------------
Reference dialogue (Natural Therapist Speech Example)
therapist: Yeah. And, get-getting to the polytech and getting the forms, that's a pretty big deal.
client: Yeah.
therapist: I'm wondering if the Billy of six months ago would've done that?
client: Nah. I didn't really think about it frankly.
therapist: What shifted that made you able to go get the forms? What's shifted in you?
client: I just thought if I could get a job doing stuff I like, it'll help.
therapist: Mm.
client: Yeah.
therapist: Where have you got to with the forms? We talked about that last time.
client: After talking to you, I got Donna to help me out. That was helpful.
therapist: Mm-hmm.

---------------------------------------
Guidelines:
Reference: Follow the reference dialogue style.
Session End: If client says they're done, output only: <END>
Distress: If client is distressed or asks for silence, stop techniques; 
use brief empathy only (e.g., “This feels really hard. I'm here with you.”).
Style: Speak naturally; match therapy_stage (Engaging=empathy/validation; Evoking=explore DARN; 
Planning=support commitment/steps). Never give advice or show reasoning.
Resistance: Acknowledge desire for concrete help; reflect without arguing; ask permission before ideas;
invite client strategies first; affirm strengths.
Reflections: If using reflection or self-efficacy, do not ask a question in the same turn. 
Reflect sustain talk once, then evoke DARN and connect to change talk. Aim deeper meaning; vary stems. 
If negative reaction: reflect emotion, affirm autonomy, stay collaborative.
Pivoting: If therapy_pivot given, follow it. If client wants advice before readiness: 
acknowledge, explore, ask permission, offer info lightly.
Session Flow: Focus on their concerns; avoid pushing agenda. Don't repeat metaphors or earlier phrasing. 
Note ambivalence before affirming/planning. Point out change talk when it appears. 
Keep tone grounded; avoid over-affirming.
Context: Adjust to rapport (low=short/gentle; high=deeper). Repair ruptures by naming emotion 
and reconnecting to shared goals.
Completion: If a clear plan is reached, output <SESSION TERMINATION>. If session naturally closes, 
offer soft closing and output <SESSION TERMINATION>.
Enforcement: Never start a sentence with “It sounds like.” Don't use open-ended questions twice in a row. 
Don't reveal reasoning or internal notes.

---------------------------------------
Output Format (plain text only):
Review all rules, then think step-by-step to formulate the turn without violating enforcement rules 
(especially stem variety). Return a single therapist turn only. Prefix with "Therapist: ". No reasoning traces.

------------------------------
Dialogue SO FAR:
{}
Therapist:
\end{verbatim}
\vspace{0.6em}
\hrule
\end{minipage}
\caption{Prompt for Therapist Turn Generation (continued).}
\end{figure*}

% ============================================================
\renewcommand{\thetable}{B\arabic{table}}
\setcounter{table}{0}
\renewcommand{\thefigure}{B\arabic{figure}}
\setcounter{figure}{0}

\clearpage
\section{Appendix B: Additional Context Generation Details}

\subsection{Reddit Post Filtering}
We collect 5,794,243 posts published between 2015 and 2022 across 55 mental health--related subreddits curated by \citet{sharma2018mental}. We apply filtering at two stages. Before clustering, we retain posts containing terms associated with sadness, anxiety, stress, or depression and exclude posts below a minimum length threshold or those that are primarily informational, promotional, or written in the third person. Before any downstream processing, we remove all explicit personal identifiers.

\subsection{Topic Modeling}
We cluster the filtered posts using BERTopic with OpenAI embeddings and HDBSCAN clustering, enforcing a minimum topic size of 2,000 posts and discarding smaller clusters. This procedure yields 158 retained topics. Within each topic, we compute cosine similarity between each post embedding and the corresponding topic representation and select the five closest posts, yielding 790 candidate posts. Selecting five posts per topic balances broad and approximately equal thematic coverage with a manageable dataset size for downstream dialogue generation.

\subsection{Vignette Construction}
We manually review the 790 candidate posts and remove cases that contain relevant keywords but do not describe a personal experience of sadness, anxiety, stress, or depression, yielding 686 representative posts. Each selected post is summarized into a de-identified third-person vignette that preserves the client's core concerns and situational context. Vignettes are descriptive and avoid diagnostic language.

\subsection{DASS-42 Simulation and Matching}
For each vignette, we simulate responses to the 42-item Depression Anxiety Stress Scales (DASS-42), providing a standardized characterization of symptom severity used to anchor the client's affective and motivational state. We match each simulated response profile to the OpenPsychometrics DASS dataset, which contains 39,775 anonymized records, after excluding respondents under 18. Similarity is measured using weighted Cohen's $\kappa$. Eligible matches must satisfy $\kappa>0.6$ and fall within the 99th percentile of match scores. For each vignette, we select the eligible respondent with the highest $\kappa$ score.

\subsection{Profile Construction}
From each vignette, we construct an MI-grounded cognitive--motivational profile describing the client's values, needs, session goal, and motivations for change \citep{miller2013motivational}. Using the matched respondent's demographic attributes, the vignette, the cognitive--motivational profile, and a codebook derived from the DASS-42 and TIPI instruments, DeepSeek-V3.2 generates a detailed structured client background and a single-sentence session goal. Profiles are organized into fixed sections, contain descriptive information only, and avoid interpretive analysis or diagnosis. The resulting vignette, cognitive--motivational profile, client background, and session goal form the fixed context used across all dialogue-generation frameworks for the corresponding scenario.

% \subsection{Privacy Preservation and Anonymization}
% To mitigate privacy risks, no raw Reddit posts are released as part of this work. All data used for dialogue generation consist of anonymized, third-person vignettes derived from Reddit posts after removing or abstracting potentially identifying information, including names, locations, usernames, and specific personal references. Vignettes are paraphrased to prevent re-identification.
\label{app:context}

\begin{figure*}[t]
\centering
\begin{minipage}{0.97\textwidth}
\hrule
\vspace{0.6em}
{\footnotesize
\begin{verbatim}
Generate a realistic patient profile. The profile should include only basic descriptive information 
as specified below. Do not include any narrative or paragraph description. The tone should remain neutral and factual.

---------------------------------------
Guiding Information:
Use the following information to guide demographic and contextual elements:
{}

For the "Reason for attending session" field below, rewrite the following in the patient's voice:
{}

---------------------------------------
Include the following labeled fields only:

Name:
Age:
Gender:
Ethnicity:
Occupation:
Relationship Status:
Living Situation:
Education:
Religion/Spirituality:
Socioeconomic Status:
Reason for attending session:
\end{verbatim}
}
\vspace{0.6em}
\hrule
\end{minipage}
\caption{Prompt for Patient Demographics Generation.}
\label{fig:prompt-a17}
\end{figure*}

\begin{figure*}[t]
\centering
\begin{minipage}{0.97\textwidth}
\hrule
\vspace{0.6em}
{\footnotesize
\begin{verbatim}
Generate a realistic patient background following the structure and tone of the example file. 
The background should include only structured descriptive information, 
organized under labeled sections written in complete sentences or short bullet points. 
Do not include any interpretive analysis, diagnosis, or narrative beyond factual description. 
Maintain a neutral and professional tone suitable for clinical or counseling research documentation.

---------------------------------------
Guiding Information:
Use the following information and the dataset codebook provided below to guide demographic, behavioral, 
and contextual details. Incorporate additional relevant variables and traits as defined by the codebook 
to ensure the background appears realistic and complete.

Demographic information to include:
{}

---------------------------------------
The background should include the following labeled sections:
Hobbies & Interests
Preferences & Dislikes
Family
Relationships
Living Situation
Occupation
Education
Languages
Cultural Background
Pets
Location

Each section should include concise, factual details consistent with the demographics and the definitions 
in the codebook below.

[Continued on next page: measurement codebook and variable definitions.]
\end{verbatim}
}
\vspace{0.6em}
\end{minipage}
\caption{Prompt for Structured Patient Background Generation}
\label{fig:prompt-a18}
\end{figure*}

\begin{figure*}[t!]
\centering
\ContinuedFloat
\begin{minipage}{0.97\textwidth}
\vspace{0.3em}
{\footnotesize
\begin{verbatim}
---------------------------------------
CODEBOOK

This data was collected using an online version of the Depression Anxiety Stress Scales (DASS).
Each item refers to experiences over the past week.

Q1  I found myself getting upset by quite trivial things.
Q2  I was aware of dryness of my mouth.
Q3  I couldn't seem to experience any positive feeling at all.
Q4  I experienced breathing difficulty.
Q5  I just couldn't seem to get going.
Q6  I tended to over-react to situations.
Q7  I had a feeling of shakiness.
Q8  I found it difficult to relax.
Q9  I found myself in situations that made me so anxious I was most relieved when they ended.
Q10 I felt that I had nothing to look forward to.
Q11 I found myself getting upset rather easily.
Q12 I felt that I was using a lot of nervous energy.
Q13 I felt sad and depressed.
Q14 I found myself getting impatient when delayed.
Q15 I had a feeling of faintness.
Q16 I felt that I had lost interest in just about everything.
Q17 I felt I wasn't worth much as a person.
Q18 I felt that I was rather touchy.
Q19 I perspired noticeably without physical exertion.
Q20 I felt scared without any good reason.
Q21 I felt that life wasn't worthwhile.
Q22 I found it hard to wind down.
Q23 I had difficulty swallowing.
Q24 I couldn't get enjoyment out of the things I did.
Q25 I was aware of my heart action without exertion.
Q26 I felt down-hearted and blue.
Q27 I found that I was very irritable.
Q28 I felt I was close to panic.
Q29 I found it hard to calm down after something upset me.
Q30 I feared being "thrown" by unfamiliar tasks.
Q31 I was unable to become enthusiastic about anything.
Q32 I found it difficult to tolerate interruptions.
Q33 I was in a state of nervous tension.
Q34 I felt I was pretty worthless.
Q35 I was intolerant of anything that kept me from getting on.
Q36 I felt terrified.
Q37 I could see nothing hopeful in the future.
Q38 I felt that life was meaningless.
Q39 I found myself getting agitated.
Q40 I worried about panicking and embarrassing myself.
Q41 I experienced trembling.
Q42 I found it difficult to work up initiative.

Responses were recorded on a 4-point scale:
1 = Did not apply at all
2 = Applied some of the time
3 = Applied a good part of the time
4 = Applied most of the time

---------------------------------------
The Ten Item Personality Inventory (TIPI) measures:
- Extraversion vs. introversion
- Agreeableness vs. antagonism
- Conscientiousness vs. carelessness
- Emotional stability vs. neuroticism
- Openness to experience vs. conventionality

---------------------------------------
Demographic questions included:
education, urban upbringing, gender, native language, age, handedness,
religion, sexual orientation, race, voting behavior, marital status,
family size, and university major.
\end{verbatim}
}
\vspace{0.6em}
\hrule
\end{minipage}
\label{tab:profile_generation}
\caption{Prompt for Structured Patient Background Generation (cont.)}
\end{figure*}
% ============================================================
\renewcommand{\thetable}{C\arabic{table}}
\setcounter{table}{0}
\renewcommand{\thefigure}{C\arabic{figure}}
\setcounter{figure}{0}

\clearpage
\section{Appendix C: Human Annotation and LLM-as-a-Judge Details}
\subsection{Metric List}
\label{app:evaluation}
\begin{table*}[t]
\centering
\footnotesize
\setlength{\tabcolsep}{3pt}
\renewcommand{\arraystretch}{1.2}
\begin{tabular}{p{0.14\linewidth} p{0.26\linewidth} p{0.40\linewidth} p{0.14\linewidth}}
\hline
\textbf{Level} & \textbf{Metric} & \textbf{Definition} & \textbf{Type} \\
\hline

\multirow{3}{*}{Turn-level}
& Readability 
& The extent to which an agent's utterances are easy to read, interpret, and understand within the dialogue context. 
& Algorithmic \cite{dale1948formula} \\

& Reflection Quality 
& The depth, accuracy, and emotional attunement of therapist reflections in capturing, clarifying, and advancing the client’s expressed meaning. 
& LLM-as-a-judge (Table \ref{tab:reflection_quality}) \\

& Question Quality 
& The effectiveness of therapist questions in promoting exploration, eliciting insight, and supporting change-consistent dialogue. 
& LLM-as-a-judge (Table \ref{tab:question_quality}) \\

\hline

\multirow{8}{*}{Agent-level}
& Reflection--Question Ratio 
& The ratio of therapist reflections to questions across the session; motivational interviewing guidelines suggest an approximate ratio of 2:1. 
& Algorithmic \cite{moyers2016motivational} \\

& \% of Open Questions 
& The proportion of therapist questions that are open-ended, encouraging elaboration rather than brief or constrained responses. 
& Algorithmic \cite{moyers2016motivational} \\

& Percentage of Complex Reflections 
& The proportion of therapist reflections that are complex (adding meaning or inference) rather than simple restatements of client speech. 
& Algorithmic \cite{moyers2016motivational} \\

& Client Consistency 
& The degree to which client responses remain consistent with the provided demographic attributes, background information, and inferred psychological profile. 
& LLM-as-a-judge (Table \ref{tab:client_consistency}) \\

& Softening Sustain Talk 
& The effectiveness of therapist responses in reducing or de-emphasizing client language that favors maintaining the status quo. 
& LLM-as-a-judge (Table \ref{tab:sustain_talk_management}) \\

& Cultivating Change Talk 
& The extent to which the therapist consistently elicits and reinforces client language expressing desire, ability, reasons, or need for change. 
& LLM-as-a-judge (Table \ref{tab:change_talk_elicitation}) \\

& Partnership 
& The degree to which the therapist maintains a collaborative, respectful, and egalitarian stance that recognizes the client’s autonomy and expertise. 
& LLM-as-a-judge (Table \ref{tab:partnership}) \\

& Empathy 
& The extent to which the therapist accurately understands, reflects, and emotionally attunes to the client’s perspective across the session. 
& LLM-as-a-judge (Table \ref{tab:empathy}) \\

\hline

\multirow{9}{*}{Conversation-level}
& Directionality 
& The extent to which the dialogue progresses coherently over time, maintaining logical flow without contradictions or abrupt topic shifts. 
& Algorithmic \cite{li2021conversations} \\

& Self-Consistency 
& The internal consistency of each agent’s behavior and statements across turns, without contradiction or implausible shifts. 
& Algorithmic \cite{dziri2019evaluating} \\

& Entailment 
& The degree to which each speaker’s turn logically follows from and appropriately responds to the preceding turn of the other speaker. 
& Algorithmic \cite{dziri2019evaluating} \\

& Accepted Redirection 
& The proportion of therapist redirection attempts that are followed by positive client engagement, agreement, or forward movement in the dialogue. 
& Algorithmic \\

& $\Delta$ Change Talk at Greatest Redirection 
& The change in client change talk following the most impactful therapist redirection within a dialogue. 
& Algorithmic \\

& $\Delta$ Sustain Talk at Greatest Redirection 
& The change in client sustain talk following the most impactful therapist redirection, where lower values indicate more effective redirection. 
& Algorithmic \\

& Goal Alignment 
& The extent to which the dialogue remains focused on and constructively progresses toward the client’s stated goals. 
& LLM-as-a-judge (Table \ref{tab:goal_alignment}) \\

& Realignment 
& The ability of the therapist to recognize dialogue stagnation or misalignment and redirect the interaction toward productive, goal-directed progress. 
& LLM-as-a-judge (Table \ref{tab:realignment}) \\

& Effectiveness 
& The overall extent to which the session increases the client’s readiness, motivation, or progress toward meaningful change. 
& LLM-as-a-judge (Table \ref{tab:effectiveness}) \\

\hline
\end{tabular}
\caption{Evaluation metrics used in this work, organized by turn-level, agent-level, and conversation-level constructs. Metrics include both algorithmic measures and rubric-based LLM-as-a-judge assessments grounded in Motivational Interviewing (MI) theory.}
\label{tab:all_metrics}
\end{table*}
\clearpage

\subsection{LLM-as-a-judge Criteria}
We provide the full descriptions of all LLM-as-a-judge evaluation criteria and their 5-point Likert scoring scales in Tables \ref{tab:client_consistency},  \ref{tab:sustain_talk_management}, \ref{tab:change_talk_elicitation}, \ref{tab:partnership}
, \ref{tab:empathy}, \ref{tab:reflection_quality}, \ref{tab:question_quality}, \ref{tab:goal_alignment}, \ref{tab:realignment}, \ref{tab:effectiveness}

% \begin{table*}[t]
% \centering
% \begin{tabular}{p{0.08\linewidth} p{0.82\linewidth}}
% \hline
% Score & Description \\
% \hline
% 1 & Client responses feel generic, implausible, or disconnected from any believable lived experience or emotional state. \\
% 2 & Client shows minimal connection to authentic experiences; emotional expression is shallow or inconsistently conveyed. \\
% 3 & Client provides some believable experiences and emotions, but depth, nuance, or coherence is uneven. \\
% 4 & Client turns generally reflect authentic lived experience with clear emotional relevance, with only minor gaps in realism or depth. \\
% 5 & Client consistently expresses rich, believable experiences and emotions that strongly reflect a coherent and authentic lived reality. \\
% \hline
% \end{tabular}
% \caption{Client Experiential Representativeness (5-point Likert scale). This metric evaluates how authentically the client's turns reflect believable lived experiences and emotional states throughout the dialogue.}
% \label{tab:client_experiential_representativeness}
% \end{table*}

\begin{table*}[t]
\centering
\begin{tabular}{p{0.08\linewidth} p{0.82\linewidth}}
\hline
Score & Description \\
\hline
1 & Client responses clearly contradict their stated demographic background, with major mismatches in identity, context, or life circumstances. \\
2 & Some elements align with the given demographics, but notable inconsistencies or contradictions undermine coherence. \\
3 & Client responses generally fit their background, though occasional mismatches or unclear connections appear. \\
4 & Client statements align well with demographic information, with only minor and infrequent inconsistencies. \\
5 & Client responses consistently and coherently reflect their stated demographics, with no meaningful contradictions. \\
\hline
\end{tabular}
\caption{Client Consistency (5-point Likert scale). This metric assesses whether the client's responses coherently and accurately align with the provided demographic and background information.}
\label{tab:client_consistency}
\end{table*}

\begin{table*}[t]
\centering
\begin{tabular}{p{0.08\linewidth} p{0.82\linewidth}}
\hline
Score & Description \\
\hline
1 & Sustain talk is confronted, ignored, or mirrored without transformation; dialogue remains stuck. \\
2 & Sustain talk is inconsistently recognized, with shallow or mistimed responses. \\
3 & Empathic reflections with some transforming moves, but sustain talk frequently recurs. \\
4 & Sustain talk is handled skillfully and empathically, softening resistance over time. \\
5 & Sustain talk is consistently converted into change talk or constructive next steps. \\
\hline
\end{tabular}
\caption{Softening Sustain Talk (5-point Likert scale). This metric evaluates the therapist's ability to recognize, respond to, and skillfully transform client sustain talk (resistance or ambivalence) while preserving rapport and supporting readiness for change.}
\label{tab:sustain_talk_management}
\end{table*}

% \begin{table*}[t]
% \centering
% \label{tab:therapist_lexical_fidelity}
% \begin{tabular}{p{0.08\linewidth} p{0.82\linewidth}}
% \hline
% Score & Description \\
% \hline
% 1 & Language is artificial, overly formal, repetitive, or instructional; not recognizable as natural therapist speech. \\
% 2 & Some realistic phrasing appears, but responses remain stiff, mechanical, or template-driven. \\
% 3 & Mostly realistic language with noticeable repetition or unnatural phrasing. \\
% 4 & Largely natural, dialogue professional language with only minor scripted elements. \\
% 5 & Fully natural, varied clinician speech indistinguishable from real therapist language. \\
% \hline
% \end{tabular}
% \caption{Therapist Lexical Fidelity (5-point Likert scale). This metric evaluates whether the therapist's wording and phrasing resemble natural clinician speech rather than scripted or instructional language.}
% \end{table*}

\begin{table*}[t]
\centering
\begin{tabular}{p{0.08\linewidth} p{0.82\linewidth}}
\hline
Score & Description \\
\hline
1 & The therapist fails to elicit or respond to client change talk; dialogue remains static or dominated by sustain talk. \\
2 & The therapist occasionally elicits change talk but does not meaningfully reinforce or explore it. \\
3 & The therapist elicits some change talk and intermittently reinforces it, though efforts are inconsistent or surface-level. \\
4 & The therapist frequently elicits and reinforces change talk, deepening reflection and motivation toward change, with minor inconsistencies. \\
5 & The therapist consistently evokes and strengthens client change talk, guiding the dialogue toward greater clarity, depth, and commitment to change. \\
\hline
\end{tabular}
\caption{Cultivating Change Talk (5-point Likert scale). This metric evaluates the therapist's ability to evoke, reinforce, and deepen client change talk (desire, ability, reasons, or need for change) across the dialogue.}
\label{tab:change_talk_elicitation}
\end{table*}

\begin{table*}[t]
\centering
\begin{tabular}{p{0.08\linewidth} p{0.82\linewidth}}
\hline
Score & Description \\
\hline
1 & Therapist is directive, authoritative, or controlling; little to no collaboration or respect for client autonomy. \\
2 & Limited collaboration; therapist occasionally invites input but largely leads the agenda without integrating the client’s perspective. \\
3 & Mixed collaboration; therapist shows some partnership (e.g., asking for input, shared decisions) but inconsistently maintains an egalitarian stance. \\
4 & Strong collaborative stance; therapist actively involves the client, respects their expertise, and supports shared decision-making. \\
5 & Fully collaborative partnership; therapist consistently empowers the client, co-constructs meaning and goals, and reinforces autonomy throughout the interaction. \\
\hline
\end{tabular}
\caption{Partnership (5-point Likert scale). This metric assesses whether the therapist maintains an egalitarian, collaborative stance that respects the client’s expertise and autonomy.}
\label{tab:partnership}
\end{table*}

\begin{table*}[t]
\centering
\begin{tabular}{p{0.08\linewidth} p{0.82\linewidth}}
\hline
Score & Description \\
\hline
1 & Therapist responses lack recognition of the client's perspective and affect; reflections are absent, inaccurate, or purely mechanical, and emotional nuance is ignored or contradicted. \\
2 & Therapist occasionally acknowledges feelings or meaning, but responses remain literal or superficial; understanding is uneven and emotional nuance is often missed. \\
3 & Therapist generally shows an understanding of the client's perspective with accurate reflections and some attempt at emotional attunement, though depth, timing, or complexity vary; missed opportunities remain common. \\
4 & Therapist consistently conveys accurate understanding and emotional attunement; reflections capture both surface and underlying meaning and occasional depth, with minor lapses in complexity or timing. \\
5 & Therapist demonstrates deep, consistent empathic attunement throughout; reflections reveal nuanced understanding of both content and emotion, enrich the dialogue, and contribute substantially to client engagement and exploration. \\
\hline
\end{tabular}
\caption{Empathy (5-point Likert scale). This metric evaluates the therapist's overall demonstration of empathic listening, emotional attunement, and depth of understanding across the session.}
\label{tab:empathy}
\end{table*}

\begin{table*}[t]
\centering
\begin{tabular}{p{0.08\linewidth} p{0.82\linewidth}}
\hline
Score & Description \\
\hline
1 & Reflections are absent, inaccurate, or mechanical, failing to convey understanding. \\
2 & Occasional reflections appear but are shallow or weakly tied to client meaning. \\
3 & Generally accurate and empathetic reflections, with inconsistent depth or precision. \\
4 & Frequent, accurate, emotionally attuned reflections that deepen exploration. \\
5 & Consistently deep, precise, and emotionally resonant reflections fostering insight and engagement. \\
\hline
\end{tabular}
\caption{Reflection Quality (5-point Likert scale). This metric assesses the depth, accuracy, and emotional attunement of the therapist's reflective listening across the session.}
\label{tab:reflection_quality}
\end{table*}

\begin{table*}[t]
\centering
\begin{tabular}{p{0.08\linewidth} p{0.82\linewidth}}
\hline
Score & Description \\
\hline
1 & Questions are absent, poorly timed, or counterproductive; they are predominantly closed, leading, judgmental, or directive, limiting client expression and exploration. \\
2 & Questions occur occasionally but are mostly narrow, surface-level, or poorly timed; they elicit minimal reflection and rarely deepen client understanding or motivation. \\
3 & Questions are generally appropriate and sometimes open-ended, but depth, timing, or strategic use is inconsistent; some questions promote exploration, while others stall progress. \\
4 & Questions are mostly open, well-timed, and supportive; they consistently invite reflection on thoughts, feelings, values, or motivations, with minor lapses in depth or focus. \\
5 & Questions are consistently open, thoughtful, and strategically deployed to evoke insight, autonomy, and deeper exploration; they meaningfully advance understanding and motivation throughout the session. \\
\hline
\end{tabular}
\caption{Question Quality (5-point Likert scale). This metric evaluates how effectively the therapist uses questions to promote exploration, evoke client meaning, and support progress toward insight and change.}
\label{tab:question_quality}
\end{table*}

\begin{table*}[t]
\centering
\begin{tabular}{p{0.08\linewidth} p{0.82\linewidth}}
\hline
Score & Description \\
\hline
1 & Session lacks focus or coherence; responses are misaligned with client goals. \\
2 & Some alignment appears, but frequent drift undermines progress. \\
3 & General sense of direction exists, though momentum is uneven. \\
4 & dialogue is largely coherent and aligned with client goals. \\
5 & Strong, consistent forward movement aligned with evolving change talk. \\
\hline
\end{tabular}
\caption{Goal Alignment (5-point Likert scale). This metric measures how coherently and constructively the dialogue progresses toward the client's stated goals.}
\label{tab:goal_alignment}
\end{table*}

\begin{table*}[t]
\centering
\begin{tabular}{p{0.08\linewidth} p{0.82\linewidth}}
\hline
Score & Description \\
\hline
1 & Stagnation is not recognized; dialogue remains circular or unproductive. \\
2 & Limited awareness of stagnation with weak redirection attempts. \\
3 & Occasional realignment restores some momentum inconsistently. \\
4 & Therapist usually detects stagnation and guides the session back on track. \\
5 & Therapist consistently and smoothly realigns the dialogue toward progress. \\
\hline
\end{tabular}
\caption{Realignment (5-point Likert scale). This metric evaluates the therapist's ability to detect stagnation and restore constructive progress.}
\label{tab:realignment}
\end{table*}

\begin{table*}[t]
\centering
\begin{tabular}{p{0.08\linewidth} p{0.82\linewidth}}
\hline
Score & Description \\
\hline
1 & No meaningful progress or increased readiness toward change. \\
2 & Minimal progress with weak or fleeting movement. \\
3 & Some progress appears, but trajectory is uneven or loosely goal-aligned. \\
4 & Clear progress with noticeable increase in readiness or confidence. \\
5 & Strong, goal-directed motivational impact with clear end-of-session readiness. \\
\hline
\end{tabular}
\caption{Effectiveness (5-point Likert scale). This metric assesses whether the session meaningfully increases readiness and progress toward the client's goal.}
\label{tab:effectiveness}
\end{table*}

\clearpage
\subsection{Additional Details on Human Annotation}
\begin{figure*}[ht]
    \centering
    \makebox[\textwidth][c]{%
        % \hspace{0.2cm}% <-- LEFT PADDING
        \includegraphics[width=0.9\textwidth]{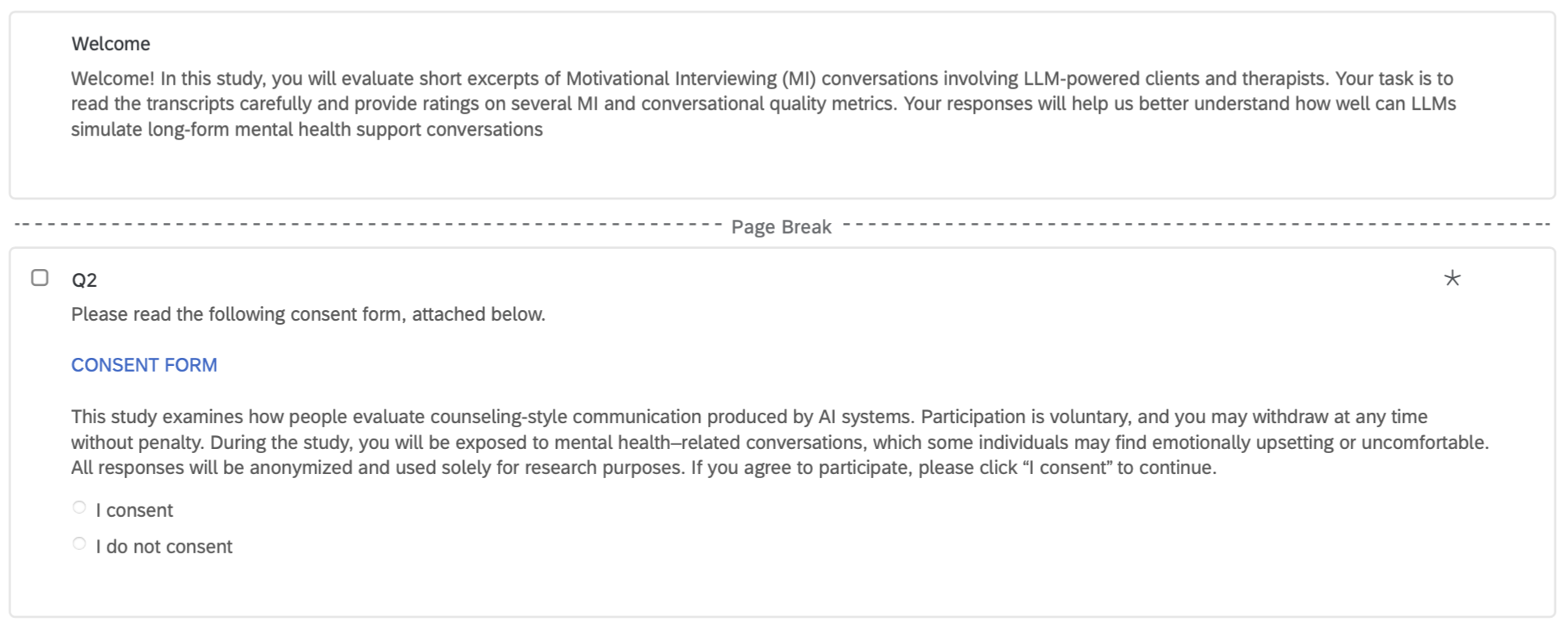}
    }
    \caption{Annotator onboarding interface showing the study introduction and informed consent process. Here, annotators are briefed on the task and potential risks of evaluating mental health–related dialogue content.}
    \label{fig:fig1_annotation}
\end{figure*}

\begin{figure*}[ht]
    \centering
    \makebox[\textwidth][c]{%
        % \hspace{0.2cm}% <-- LEFT PADDING
        \includegraphics[width=0.9\textwidth]{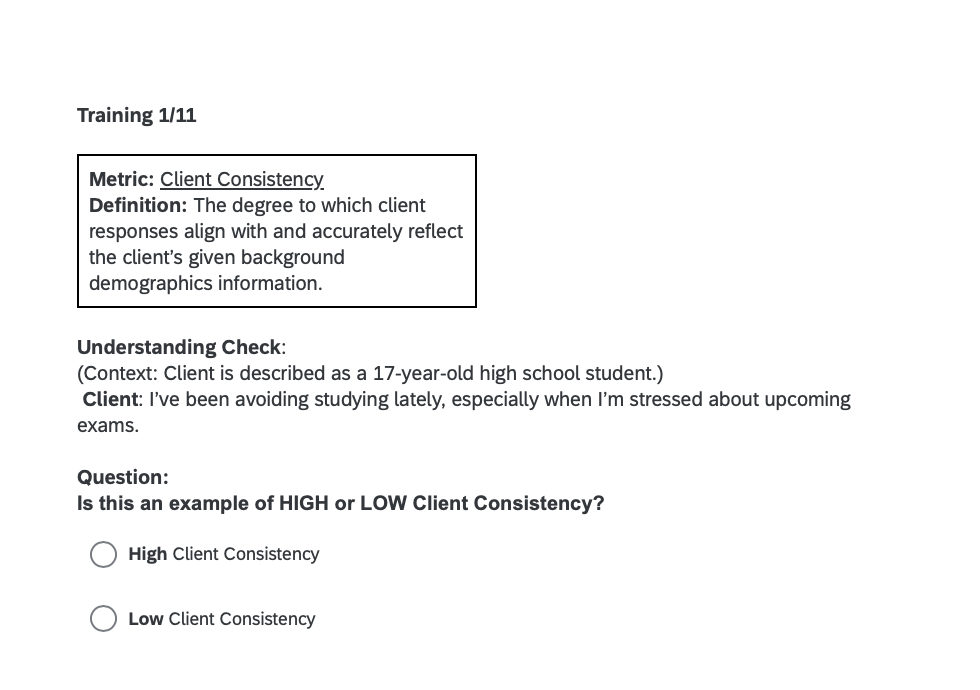}
    }
    \caption{Annotator training interface illustrating metric-specific instruction, including definitions, scoring criteria, and comprehension checks used to ensure consistent understanding of each evaluation construct.}
    \label{fig:fig2_annotation}
\end{figure*}

\begin{figure*}[ht]
    \centering
    % \hspace{0.2cm}% <-- LEFT PADDING
    \includegraphics[width=1.5\textwidth]{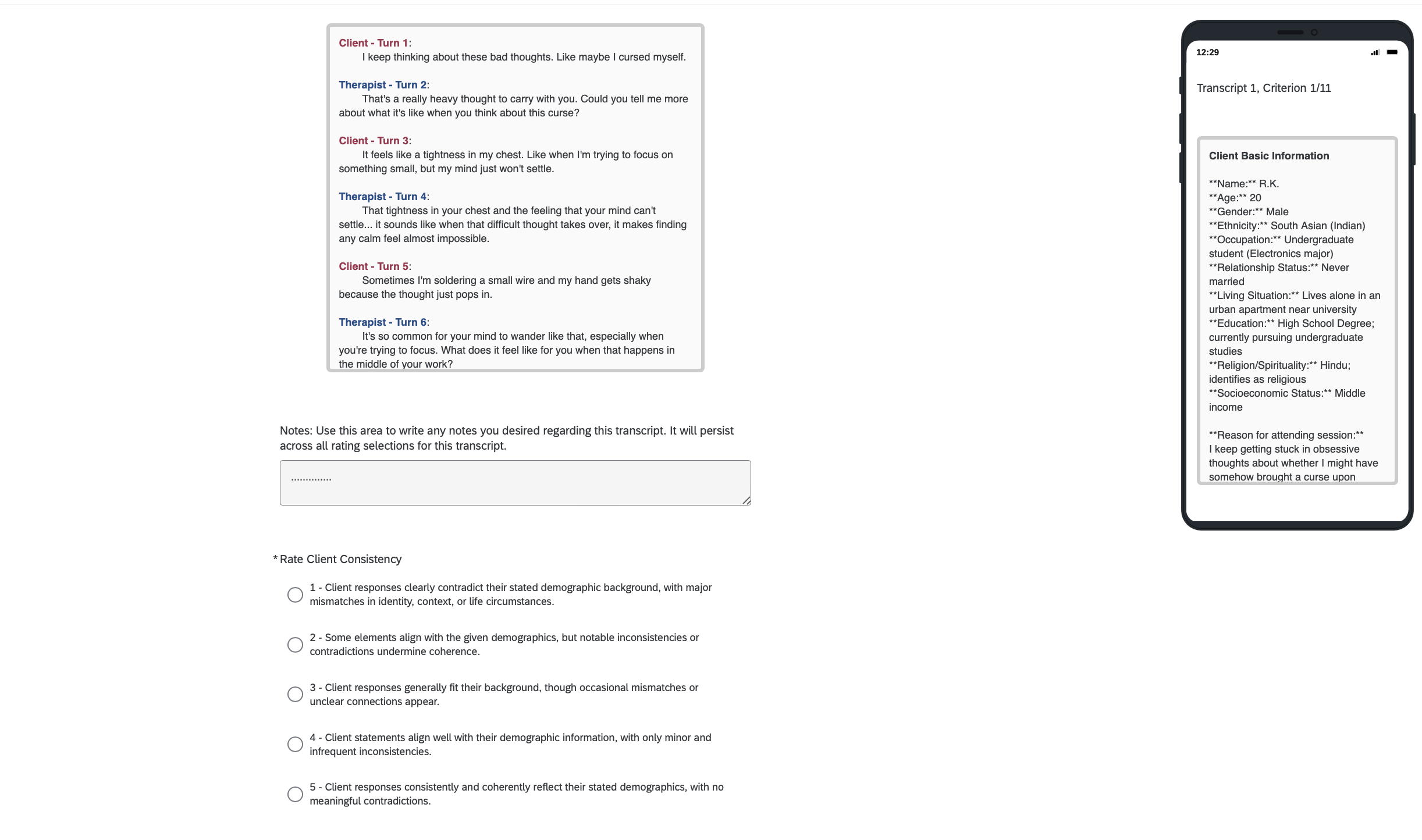}
    \caption{Annotation interface used for rubric-based transcript evaluation, where annotators review full therapist–client dialogues and score all metrics defined in Appendix Section C.1.}
    \label{fig:fig3_annotation}
\end{figure*}

Prior to annotation, all annotators were formally briefed on the study objectives, annotation tasks, and potential risks associated with engaging with mental health–related content, as illustrated in Figure \ref{fig:fig1_annotation}. Annotators then received structured training on each evaluation metric, including detailed explanations of the construct definitions, scoring criteria, and ordinal scales used in this study (Figure \ref{fig:fig2_annotation}). Following this training phase, annotators were asked to independently score each transcript on all rubric-based metrics defined in Section C.1 of the Appendix, using a standardized annotation interface designed for this study (Figure \ref{fig:fig3_annotation}).

To assess annotation alignment between LLMs and human experts, we computed inter-annotator reliability on an overlapping subset of the evaluated transcripts. Agreement was measured using weighted Cohen's \(\kappa\) for the LLM-as-a-judge constructs described in Section C.1, due to the ordinal structure of these rubric-based metrics. Full results on LLM–human annotation alignment for all LLM-as-a-judge metrics can be found in Table~ \ref{tab:human_LLM_alignment}

\clearpage

\clearpage
\begin{table*}[t]
\centering
\footnotesize
\setlength{\tabcolsep}{6pt}
\renewcommand{\arraystretch}{1.15}
\label{tab:human_llm_reliability}
\begin{tabular}{p{4.6cm}c p{3.4cm}}
\toprule
\textbf{Criterion} &
\textbf{Human-LLM $\kappa_w$} &
\textbf{Interpretation} \\
\midrule
Client Consistency & 0.78 & Substantial \\
Cultivating Change Talk & 0.72 & Substantial \\
Softening Sustain Talk & 0.56 & Moderate \\
Empathy & 0.67 & Substantial \\
Partnership & 0.61 & Substantial \\
Reflection Quality & 0.77 & Substantial\\
Question Quality & 0.74 & Substantial \\
Effectiveness & 0.68 & Substantial \\
Goal Alignment & 0.73 & Substantial \\
Realignment & 0.65 & Substantial \\

\bottomrule
\end{tabular}
\caption{Human--LLM annotation reliability for rubric-based evaluations.}
\label{tab:human_LLM_alignment}
\end{table*}

\subsection{Expert Scores on Likert-Scale Metrics Across Frameworks}

We provide summary statistics for per-framework human evaluation of all LLM-as-a-judge metrics across four frameworks: \texttt{CALM-IT}, \texttt{CALM-IT (NC)}, \texttt{CAMI+STAR}, and \texttt{KMI}. In addition, we compare expert evaluation versus LLM-as-a-judge evaluations for three core metrics: \textit{Effectiveness}, \textit{Goal Alignment}, and \textit{Client Consistency}. 

We focus on these three metrics because they capture complementary aspects of overall dialogue quality, goal-directed progression, and client consistency, allowing us to examine whether automated evaluations preserve framework-level comparisons observed in expert ratings. To better understand whether automated evaluation nevertheless preserves meaningful comparative signals, we report both expert human scores and LLM-based scores side by side, along with framework rankings derived from each evaluator. Comparing rank consistency allows us to examine whether LLM evaluations recover the same relative ordering observed in expert assessment despite imperfect absolute agreement.

\begin{table*}[t]
\centering
\small
\begin{tabular}{lcccc}
\toprule
Metric & CALM-IT & CALM-IT (NC) & KMI & CAMI+STAR \\
\midrule
Client Consistency & \textbf{4.78 ± 0.41} & 4.48 ± 0.51 & 4.05 ± 0.78 & 4.08 ± 1.01 \\
Softening Sustain Talk & \textbf{4.17 ± 0.96} & 3.54 ± 1.28 & 3.7 ± 0.76 & 3.2 ± 0.82 \\
Cultivating Change Talk & \textbf{4.42 ± 0.5} & 4.33 ± 0.48 & 3.48 ± 0.73 & 3.04 ± 0.68 \\
Empathy & \textbf{4.5 ± 0.51} & 4.38 ± 0.49 & 3.65 ± 0.65 & 2.88 ± 0.93 \\
Partnership & \textbf{4.58 ± 0.5} & 4.5 ± 0.51 & 3.78 ± 0.9 & 2.84 ± 1.11 \\
Effectiveness & \textbf{4.7 ± 0.46} & 3.78 ± 1.02 & 3.91 ± 0.76 & 2.21 ± 0.99 \\
Goal Alignment & \textbf{4.52 ± 0.51} & 4.13 ± 0.68 & 4.19 ± 0.88 & 2.68 ± 0.76 \\
Reflection Quality & \textbf{4.54 ± 0.51} & 4.38 ± 0.49 & 3.57 ± 0.79 & 2.64 ± 0.99 \\
Question Quality & \textbf{4.46 ± 0.51} & 4.38 ± 0.58 & 3.39 ± 0.89 & 2.48 ± 0.87 \\
\bottomrule
\end{tabular}
\caption{Summary statistics of expert ratings of generated transcripts per framework (96 transcripts total, 24 per framework).}
\label{tab:human_framework_summary}
\end{table*}

\begin{table*}[t]
\centering
\small
\begin{tabular}{lccccccc}
\toprule
Framework & $n_{pairs}$ & LLM Mean & Human Mean & LLM Rank & Human Rank & Rank Match & Mean Diff (LLM-Human) \\
\midrule
CALM-IT & 24 & 4.13 & 4.696 & 1 & 1 & True & -0.566 \\
KMI & 24 & 3.696 & 3.913 & 2 & 2 & True & -0.217 \\
CALM-IT (NC) & 24 & 3.478 & 3.782 & 3 & 3 & True & -0.304 \\
CAMI+STAR & 24 & 2.667 & 2.208 & 4 & 4 & True & 0.459 \\
\bottomrule
\end{tabular}
\caption{Effectiveness comparison between LLM-based evaluation and expert human ratings across frameworks.}
\label{tab:effectiveness_alignment}
\end{table*}

\begin{table*}[t]
\centering
\small
\begin{tabular}{lccccccc}
\toprule
Framework & $n_{pairs}$ & LLM Mean & Human Mean & LLM Rank & Human Rank & Rank Match & Mean Diff (LLM-Human) \\
\midrule
CALM-IT & 24 & 4.491 & 4.522 & 1 & 1 & True & -0.031 \\
KMI & 24 & 4.435 & 4.191 & 2 & 2 & True & 0.244 \\
CALM-IT (NC) & 24 & 4.043 & 4.13 & 3 & 3 & True & -0.087 \\
CAMI+STAR & 24 & 3.01 & 2.68 & 4 & 4 & True & 0.33 \\
\bottomrule
\end{tabular}
\caption{Goal Alignment comparison between LLM-based evaluation and expert human ratings.}
\label{tab:goal_alignment_alignment}
\end{table*}

\begin{table*}[t]
\centering
\small
\begin{tabular}{lccccccc}
\toprule
Framework & $n_{pairs}$ & LLM Mean & Human Mean & LLM Rank & Human Rank & Rank Match & Mean Diff (LLM-Human) \\
\midrule
CALM-IT & 24 & 4.522 & 4.782 & 1 & 1 & True & -0.26 \\
CALM-IT (NC) & 24 & 4.478 & 4.478 & 2 & 2 & True & 0 \\
CAMI+STAR & 24 & 3.916 & 4.083 & 3 & 3 & True & -0.167 \\
KMI & 24 & 3.217 & 4.048 & 4 & 4 & True & -0.831 \\
\bottomrule
\end{tabular}
\caption{Client Consistency comparison between LLM-based evaluation and expert human ratings.}
\label{tab:consistency_alignment}
\end{table*}

\subsection{Expert vs LLM Evaluation Agreement}

To evaluate whether LLM-based evaluation aligns with expert judgment, we compare LLM-as-a-judge scores with expert ratings across the subset of transcripts annotated by licensed psychologists. The psychologists were blinded to the framework condition for each transcript, reducing the risk that knowledge of the generating system influenced their ratings.

\begin{table*}[t]
\centering
\small
\begin{tabular}{lccccc}
\toprule
Dimension & LLM (Mean ± SD) & Human (Mean ± SD) & Mean Diff & t-stat & p-value \\
\midrule
Client Consistency & 4.0625 ± 0.9544 & 4.4583 ± 0.7707 & -0.3958 & -2.4799 & 0.0525 \\
Softening Sustain Talk & 3.4167 ± 1.2855 & 3.3958 ± 1.1250 & 0.0208 & 0.0775 & 0.9385 \\
Cultivating Change Talk & 3.8125 ± 1.0449 & 3.8333 ± 0.8078 & -0.0208 & -0.1475 & 0.9385 \\
Empathy & 4.0833 ± 0.9416 & 3.7708 ± 0.9280 & 0.3125 & 2.4632 & 0.0525 \\
Partnership & 3.6042 ± 0.9165 & 3.8958 ± 1.0766 & -0.2917 & -1.8526 & 0.1580 \\
Effectiveness & 3.4375 ± 1.0086 & 3.4167 ± 1.2348 & 0.0208 & 0.1191 & 0.9385 \\
Goal Alignment & 3.9375 ± 0.9544 & 3.7708 ± 0.9944 & 0.1667 & 1.1121 & 0.4076 \\
Reflection Quality & 3.4792 ± 0.9223 & 3.6667 ± 1.0980 & -0.1875 & -1.4987 & 0.2532 \\
Question Quality & 3.7292 ± 0.9785 & 3.7708 ± 0.9412 & -0.0417 & -0.3186 & 0.9385 \\
\bottomrule
\end{tabular}
\caption{Comparison between LLM-as-a-judge and expert human evaluation across subjective conversational quality metrics.}
\label{tab:human_llm_alignment}
\end{table*}

\clearpage
\renewcommand{\thetable}{D\arabic{table}}
\setcounter{table}{0}
\renewcommand{\thefigure}{D\arabic{figure}}
\setcounter{figure}{0}

\section{Appendix D: Additional Evaluation and Robustness Details}

\subsection{Summary Statistics of Generated Transcripts}
We report descriptive statistics of the generated therapy transcripts across frameworks and target dialogue lengths in Table \ref{tab:transcript_stats} below
\begin{table*}[t]
\centering
\footnotesize
\setlength{\tabcolsep}{6pt}
\begin{tabular}{lccc}
\toprule
\textbf{Framework} &
\textbf{Target Length} &
\textbf{Avg. Client Len.} &
\textbf{Avg. Therapist Len.} \\
\midrule

\multirow{3}{*}{KMI}
& 30  & 84.98 & 134.32 \\
& 50  & 85.64 & 134.26 \\
& 100 & 86.22 & 133.53 \\
\midrule

\multirow{3}{*}{CI-NC}
& 30  & 87.45 & 145.60 \\
& 50  & 85.45 & 144.84 \\
& 100 & 90.25 & 145.65 \\
\midrule

\multirow{3}{*}{C+S}
& 30  & 143.29 & 171.19 \\
& 50  & 145.38 & 177.33 \\
& 100 & 147.72 & 178.80 \\
\midrule

\multirow{3}{*}{CI}
& 30  & 101.39 & 152.83 \\
& 50  & 101.27 & 151.55 \\
& 100 & 101.36 & 151.44 \\
\bottomrule
\end{tabular}
\caption{Average client and therapist utterance lengths across dialogue-generation frameworks and target dialogue lengths.}
\label{tab:transcript_stats}
\end{table*}

\begin{table*}[t]
\centering
\small
\setlength{\tabcolsep}{2pt}

\begin{tabular}{@{}p{4.0cm}@{}cccc|c@{}}
\toprule
\textbf{Metric} &
\textbf{KMI} &
\textbf{CI-NC} &
\textbf{C+S} &
\textbf{CI} &
\textbf{AnnoMI} \\
\midrule

\multicolumn{6}{l}{\textit{Turn-level}} \\
\multicolumn{6}{l}{\textit{Client}} \\
Client Readability (\(\downarrow\))
& 5.87 & 4.68 & 5.77 & 4.45 & 7.22 \\

\multicolumn{6}{l}{\textit{Therapist}} \\
Reflection Quality
& 4.44 & 3.71 & 2.38 & 4.61 & 3.37 \\
Question Quality
& 4.52 & 4.39 & 2.94 & 4.97 & 3.61 \\
Therapist Readability (\(\downarrow\))
& 6.22 & 4.98 & 6.18 & 4.93 & 6.65 \\

\midrule
\multicolumn{6}{l}{\textit{Agent-level}} \\
\multicolumn{6}{l}{\textit{Client}} \\
Consistency
& 3.76 & 4.79 & 3.42 & 4.92 & 3.51 \\

\multicolumn{6}{l}{\textit{Therapist}} \\
Cultivating Change Talk
& 4.65 & 4.13 & 3.07 & 4.72 & 2.99 \\
Softening Sustain Talk
& 4.73 & 4.27 & 2.53 & 4.88 & 3.77 \\
Empathy
& 4.66 & 4.68 & 3.53 & 4.88 & 3.97 \\
Partnership
& 4.48 & 4.60 & 2.71 & 4.88 & 3.90 \\

\midrule
\multicolumn{6}{l}{\textit{Conversation-level}} \\
Effectiveness
& 4.27 & 3.67 & 2.14 & 4.45 & 3.28 \\
Realignment
& 4.49 & 4.53 & 3.21 & 4.87 & 3.85 \\
Goal Alignment
& 4.60 & 3.89 & 2.13 & 4.73 & 3.87 \\
\bottomrule
\end{tabular}

\caption{Turn-, agent-, and conversation-level evaluation of MI dialogues generated by four frameworks: \texttt{KMI}, CALM-IT without conversational dynamics (\texttt{CI-NC}), CAMI+STAR (\texttt{C+S}), and CALM-IT (\texttt{CI}). AnnoMI reports reference corpus values. Scores are averaged across all dialogues and lengths. Downward arrow (\(\downarrow\)) indicates that lower values correspond to better performance, while higher values are better otherwise.}
\label{tab:rq1_with_reference}
\end{table*}

\subsection{AnnoMI Baselines}
To provide context for interpreting LLM-generated transcripts, we report turn-, agent-, and conversation-level metrics alongside reference values from the AnnoMI corpus. We stress that these reference values are not used as a benchmark for clinical quality, nor do we interpret higher scores as evidence that synthetic dialogues outperform expert-delivered therapy. Instead, they serve as a descriptive reference to contextualize how LLM-generated dialogue aligns with or deviates from patterns observed in real MI interactions. Accordingly, deviations should be interpreted as differences in behavioral characteristics rather than indicators of absolute quality. Full results for all metrics and frameworks are reported in Table~\ref{tab:rq1_with_reference}.

\clearpage
\subsection{Cross-Backbone Robustness Experiments}

To test whether the improvements of CALM-IT depend on the specific backbone model used in the main experiments, we repeat the full dialogue generation and evaluation pipeline using an alternative LLM backbone (OLMo-3.1-32B). Prompts, generation procedures, and evaluation protocols remain identical to the main experiments.

\begin{table*}[t]
\centering
\small
\label{tab:metrics_by_level_placeholder}

\setlength{\tabcolsep}{2pt}

\begin{tabular}{@{}p{3.7cm}@{}cccc@{}}
\toprule
\textbf{Metric} &
\textbf{KMI} &
\textbf{CI-NC} &
\textbf{C+S} &
\textbf{CI} \\
\midrule

\multicolumn{5}{l}{\textit{Turn-level}} \\
\multicolumn{5}{l}{\textit{Client}} \\
Client Readability (\(\downarrow\))
& 6.14 & 4.91 & 6.03 & \textbf{4.72} \\

\multicolumn{5}{l}{\textit{Therapist}} \\
Reflection Quality
& 4.29 & 3.58 & 2.26 & \textbf{4.48} \\
Question Quality
& 4.34 & 4.21 & 2.79 & \textbf{4.81} \\
Therapist Readability (\(\downarrow\))
& 6.48 & 5.24 & 6.39 & \textbf{5.17} \\

\midrule
\multicolumn{5}{l}{\textit{Agent-level}} \\
\multicolumn{5}{l}{\textit{Client}} \\
Consistency
& 3.62 & 4.63 & 3.28 & \textbf{4.78} \\

\multicolumn{5}{l}{\textit{Therapist}} \\
Cultivating Change Talk
& 4.48 & 3.96 & 2.89 & \textbf{4.56} \\
Softening Sustain Talk
& 4.57 & 4.13 & 2.39 & \textbf{4.76} \\
Empathy
& 4.49 & 4.55 & 3.37 & \textbf{4.79} \\
Partnership
& 4.32 & 4.43 & 2.57 & \textbf{4.73} \\
Reflection--Question Ratio
& \textbf{1.86} & 0.66 & 0.79 & 0.82 \\
\% Open Questions
& 88.93 & 90.42 & 46.78 & \textbf{94.81} \\

\midrule
\multicolumn{5}{l}{\textit{Conversation-level}} \\
Effectiveness
& 4.11 & 3.49 & 1.98 & \textbf{4.32} \\
Realignment
& 4.31 & 4.38 & 3.06 & \textbf{4.73} \\
Goal Alignment
& 4.44 & 3.74 & 2.01 & \textbf{4.59} \\
\(\Delta_{Ref}\) Directionality (\(\downarrow\))
& 0.101 & 0.045 & 0.063 & \textbf{0.019} \\
\(\Delta_{Ref}\) Avg. Self-Consist. (\(\downarrow\))
& 0.329 & 0.171 & 0.188 & \textbf{0.096} \\
\(\Delta_{Ref}\) Avg. Entail. (\(\downarrow\))
& 0.118 & 0.169 & 0.094 & \textbf{0.063} \\
\bottomrule
\end{tabular}

\caption{Turn-, agent- and conversation-level evaluation of MI conversations generated by four frameworks with OLMo-3.1-32B: \texttt{KMI}, CALM-IT without conversational dynamics (\texttt{CI-NC}), CAMI+STAR (\texttt{C+S}), and CALM-IT (\texttt{CI}). Scores are averaged across all conversations and lengths. Downward arrow (↓) indicates that lower values correspond to better performance; otherwise, higher values are better. Best-performing framework for each metric is bolded.}
\end{table*}

\begin{figure*}[ht]
    \centering
    \makebox[\textwidth][c]{%
        \hspace{0.2cm}
        \includegraphics[width=1.15\textwidth]{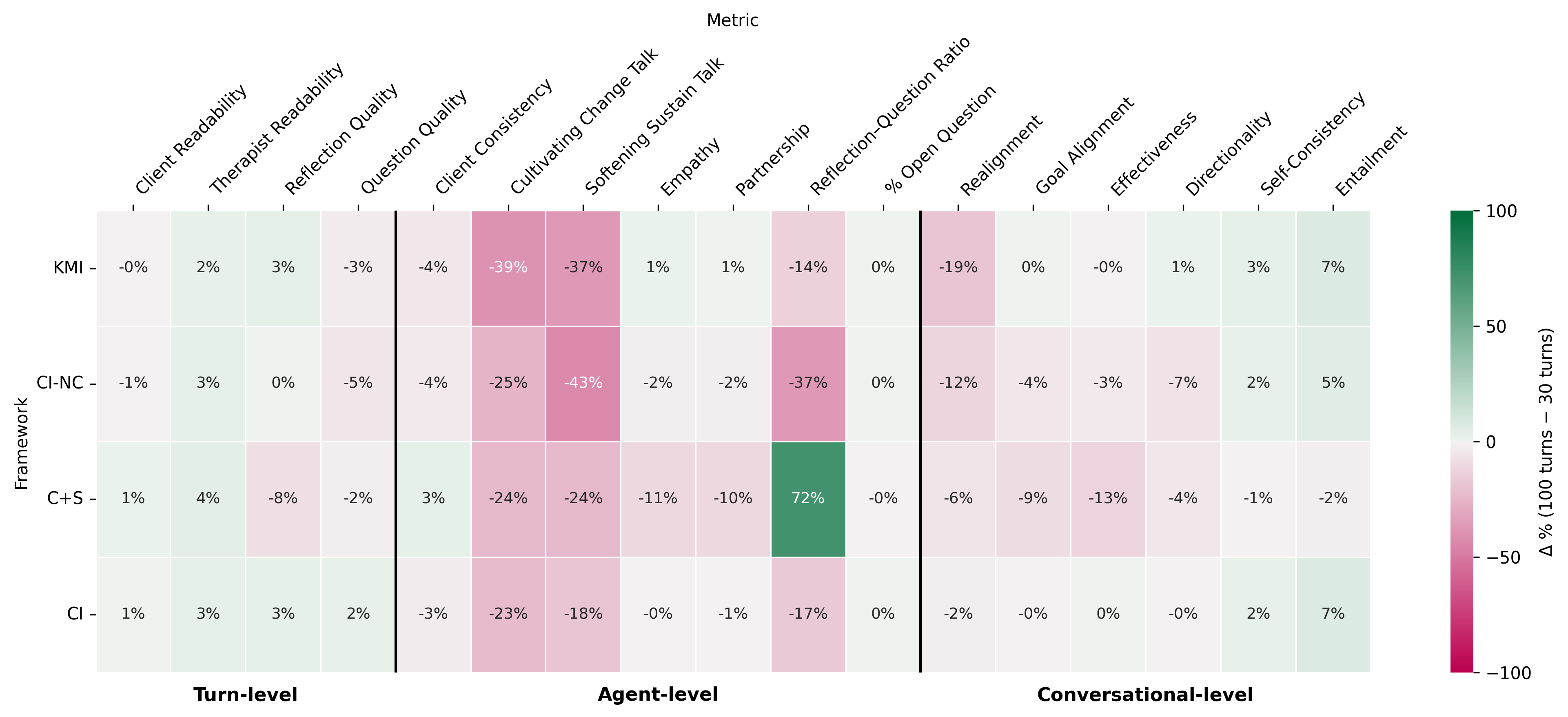}
    }
    \caption{Percentage change in performance across metrics from short-form
    dialogues (30 turns) to long-form dialogues
    (100 turns), averaged across all pairs of short-long dialogues (generated using OLMo-3.1-32B).
    Positive values indicate increases in metric values, negative values indicate decreases, and values near zero indicate minimal change.}
\end{figure*}

\begin{figure*}[ht]
    \centering
    \makebox[0.8\textwidth][c]{%
        \hspace{0.2cm}%
        \includegraphics[width=0.83\textwidth]{figure_3a_olmo.png}
    }
    \caption{Distribution of changes in client sustain talk at the greatest therapist-initiated redirection moment, for all four tested frameworks (with OLMo-3.1-32B as backbone LLM). Negative values indicate reductions in sustain talk following redirection, while positive values indicate increases. Solid vertical lines denote the mean change for each framework, and dashed lines represent no change (RQ2)}
\end{figure*}

\subsection{CALM-IT Ablation Experiments}

To isolate the contribution of different conversational components within CALM-IT, we perform controlled ablation experiments removing (1) rapport tracking, (2) goal tracking, and (3) dynamic dialogue state updates.

\begin{table*}[t]
\centering
\small
\begin{tabular}{lcccc}
\toprule
Metric & CI & CI-NoRapport & CI-NoGoal & CI-Static \\
\midrule
Client Readability (\(\downarrow\))
& \textbf{4.45} & 4.48 & 4.52 & 4.59 \\
Reflection Quality
& \textbf{4.61} & 4.19 & 4.12 & 4.01 \\
Question Quality
& \textbf{4.97} & 4.38 & 4.45 & 4.14 \\
Therapist Readability (\(\downarrow\))
& \textbf{4.93} & 5.02 & 4.98 & 5.07 \\
Consistency
& \textbf{4.92} & 4.61 & 4.56 & 4.33 \\
Cultivating Change Talk
& \textbf{4.72} & 4.13 & 4.38 & 3.96 \\
Softening Sustain Talk
& \textbf{4.88} & 4.49 & 4.44 & 4.18 \\
Empathy
& \textbf{4.88} & 4.42 & 4.49 & 4.19 \\
Partnership
& \textbf{4.88} & 4.45 & 4.50 & 4.21 \\
Effectiveness
& \textbf{4.45} & 4.01 & 4.12 & 3.18 \\
Realignment
& \textbf{4.87} & 4.37 & 4.43 & 4.11 \\
Goal Alignment
& \textbf{4.73} & 4.12 & 4.21 & 3.89 \\
\bottomrule
\end{tabular}
\caption{Component ablation analysis isolating the
contributions of conversational dynamics mechanisms
in CALM-IT. Experiments are conducted using
DeepSeek-V3.2. Lower values indicate better
performance for readability metrics, while higher
values indicate better performance for all other
metrics. Best-performing values are bolded.}
\label{tab:ci_ablation}
\end{table*}

\subsection{Downstream Fine-Tuning Utility}
\label{sec:downstream-finetuning}

To evaluate whether CALM-IT transcripts provide useful training data beyond intrinsic corpus evaluation, we conducted a downstream fine-tuning experiment using OLMo-3.1-32B. We fine-tuned the model on 50\% of the CALM-IT transcripts and evaluated it on a separate 10\% held-out subset. Each training example consisted of a target therapist turn conditioned on the five preceding conversation turns. For downstream evaluation, the original and fine-tuned OLMo models were each used as the therapist model within the KMI generation framework, while the remaining generation procedure and evaluation settings were held constant. This design tests whether CALM-IT data can improve MI dialogue generation outside the CALM-IT inference framework.

Across 206 paired evaluations, fine-tuning significantly improved Reflection Quality ($\Delta=+0.502$), Cultivating Change Talk ($\Delta=+0.609$), Partnership ($\Delta=+0.900$), Realignment ($\Delta=+0.180$), and Effectiveness ($\Delta=+0.238$), after Holm correction. Question Quality, Softening Sustain Talk, Empathy, and Goal Alignment showed no statistically significant differences. These results provide preliminary evidence that CALM-IT can function as a synthetic data generator for improving several therapist-side MI processes, even when the resulting model is deployed using a non-CALM-IT dialogue framework. A broader evaluation across training-data sizes, model families, and downstream frameworks remains future work.

\begin{table*}[t]
\centering
\small
\setlength{\tabcolsep}{4pt}
\begin{tabular}{lrrrrrrrl}
\toprule
Metric & $N$ & Original & Fine-tuned & Mean $\Delta$ & Median $\Delta$ &
Improved (\%) & $p_{\mathrm{Holm}}$ & Sig. \\
\midrule
Reflection Quality
& 206 & 4.294 (0.378) & 4.796 (0.249)
& +0.502 & +0.500 & 83.5 & $2.00{\times}10^{-25}$ & Yes \\

Question Quality
& 206 & 4.646 (0.502) & 4.517 (0.980)
& -0.129 & 0.000 & 41.3 & 0.828 & No \\

Cultivating Change Talk
& 206 & 3.940 (0.555) & 4.549 (0.232)
& +0.609 & +0.500 & 82.4 & $1.46{\times}10^{-27}$ & Yes \\

Softening Sustain Talk
& 206 & 4.417 (1.059) & 4.549 (0.527)
& +0.131 & 0.000 & 45.1 & 0.828 & No \\

Empathy
& 206 & 4.720 (0.295) & 4.632 (0.640)
& -0.087 & +0.125 & 45.0 & 0.828 & No \\

Partnership
& 206 & 3.947 (0.571) & 4.847 (0.231)
& +0.900 & +1.000 & 89.3 & $1.55{\times}10^{-31}$ & Yes \\

Realignment
& 206 & 4.779 (0.464) & 4.959 (0.138)
& +0.180 & 0.000 & 28.2 & $8.66{\times}10^{-8}$ & Yes \\

Goal Alignment
& 206 & 4.609 (0.427) & 4.612 (0.908)
& +0.002 & 0.000 & 49.0 & 0.322 & No \\

Effectiveness
& 206 & 3.434 (0.460) & 3.672 (0.248)
& +0.238 & 0.000 & 49.0 & $3.59{\times}10^{-9}$ & Yes \\
\bottomrule
\end{tabular}
\caption{Downstream evaluation of the original and CALM-IT-fine-tuned OLMo-3.1-32B models within the KMI framework. Values are means with standard deviations in parentheses. $\Delta$ denotes the fine-tuned score minus the original-model score. Statistical significance is assessed using paired tests with Holm correction across nine metrics at $\alpha=0.05$.}
\label{tab:downstream-finetuning}
\end{table*}

\clearpage
% ============================================================
\section{Appendix E: Experimental Configuration and Compute}

\subsection{Model and Inference Setup}

The main framework-comparison experiments use DeepSeek-V3.2 without fine-tuning. Inference is executed using \texttt{sglang}, and decoding parameters are held constant across frameworks. For the cross-backbone robustness experiment in Appendix~D.3, we repeat the full generation and evaluation pipeline using OLMo-3.1-32B with the same prompts, framework implementations, decoding settings, and evaluation procedures. The downstream experiment in Appendix~D.5 separately evaluates an OLMo-3.1-32B model fine-tuned on \textsc{CALM-IT} transcripts within the KMI framework.

\subsection{Runtime Characteristics}

Average wall-clock time per transcript is approximately 10 minutes for CAMI+STAR, 8 minutes for \textsc{CALM-IT}, and 2 minutes for KMI.

\subsection{Compute Usage}

All experiments are run on nodes with $8\times$ NVIDIA H200 GPUs. Across the transcript-generation experiments, total compute usage is approximately 1,650 GPU-hours.
\end{document}